\documentclass[sigconf]{acmart}

\usepackage{enumitem}
\usepackage{amsmath}
\usepackage{bm}
\usepackage{multirow}
\usepackage{subfigure}
\usepackage{algorithm}
\usepackage{algorithmic}
\usepackage{dsfont}

\newcommand\red[1]{\textcolor{red}{#1}}
\newcommand\blue[1]{\textcolor{blue}{#1}}

\AtBeginDocument{%
  }

\copyrightyear{2025}
\acmYear{2025}
\setcopyright{acmlicensed}\acmConference[SIGSPATIAL '25]{The 33rd ACM
International Conference on Advances in Geographic Information
Systems}{November 3--6, 2025}{Minneapolis, MN, USA}
\acmBooktitle{The 33rd ACM International Conference on Advances in
Geographic Information Systems (SIGSPATIAL '25), November 3--6, 2025,
Minneapolis, MN, USA}
\acmDOI{10.1145/3748636.3762709}
\acmISBN{979-8-4007-2086-4/2025/11}

\begin{document}

\title[UQGNN: Uncertainty Quantification of Graph Neural Networks]{UQGNN: Uncertainty Quantification of Graph Neural Networks for Multivariate Spatiotemporal Prediction}

\author{Dahai Yu}
\affiliation{
  \institution{Florida State University}
  \city{Tallahassee, Florida}
  \country{USA}
}
\email{dahai.yu@fsu.edu}

\author{Dingyi Zhuang}
\affiliation{
  \institution{Massachusetts Institute of Technology}
  \city{Cambridge, Massachusetts}
  \country{USA}
}
\email{dingyi@mit.edu}

\author{Lin Jiang}
\affiliation{
  \institution{Florida State University}
  \city{Tallahassee, Florida}
  \country{USA}
}
\email{lin.jiang@fsu.edu}

\author{Rongchao Xu}
\affiliation{
  \institution{Florida State University}
  \city{Tallahassee, Florida}
  \country{USA}
}
\email{rx21a@fsu.edu}

\author{Xinyue Ye}
\affiliation{
  \institution{University of Alabama, Tuscaloosa}
  \city{Tuscaloosa, Alabama}
  \country{USA}
}
\email{xye10@ua.edu}

\author{Yuheng Bu}
\affiliation{
  \institution{UC Santa Barbara}
  \city{Santa Barbara, California}
  \country{USA}
}
\email{buyuheng@ucsb.edu}

\author{Shenhao Wang}
\affiliation{
  \institution{University of Florida}
  \city{Gainesville, Florida}
  \country{USA}
}
\email{shenhaowang@ufl.edu}

\author{Guang Wang}
\authornote{Prof. Guang Wang is the corresponding author.}
\affiliation{
  \institution{Florida State University}
  \city{Tallahassee, Florida}
  \country{USA}
}
\email{guang@cs.fsu.edu}

\renewcommand{\shortauthors}{Yu et al.}

\begin{abstract}
Spatiotemporal prediction plays a critical role in numerous real-world applications such as urban planning, transportation optimization, disaster response, and pandemic control.
In recent years, researchers have made significant progress by developing advanced deep learning models for spatiotemporal prediction. However, most existing models are deterministic, i.e., predicting only the expected mean values without quantifying uncertainty, leading to potentially unreliable and inaccurate outcomes.
While recent studies have introduced probabilistic models to quantify uncertainty, they typically focus on a single phenomenon (e.g., taxi, bike, crime, or traffic crashes), thereby neglecting the inherent correlations among heterogeneous urban phenomena. 
To address the research gap, we propose a novel Graph Neural Network with Uncertainty Quantification, termed UQGNN for multivariate spatiotemporal prediction. UQGNN introduces two key innovations: (i) an Interaction-aware Spatiotemporal Embedding Module that integrates a multivariate diffusion graph convolutional network and an interaction-aware temporal convolutional network to effectively capture complex spatial and temporal interaction patterns, and (ii) a multivariate probabilistic prediction module designed to estimate both expected mean values and associated uncertainties.
Extensive experiments on four real-world multivariate spatiotemporal datasets from Shenzhen, New York City, and Chicago demonstrate that UQGNN consistently outperforms state-of-the-art baselines in both prediction accuracy and uncertainty quantification. For example, on the Shenzhen dataset, UQGNN achieves a 5\% improvement in both prediction accuracy and uncertainty quantification.

\end{abstract}





\maketitle

\section{Introduction}\label{introduction}

Spatiotemporal prediction has garnered significant attention from both academic and industry communities due to its critical role in a wide range of real-world applications, such as urban planning~\cite{jiang2018deepurbanmomentum, richly2022budget, li2022efficient, shi2024data, ye2025urban}, epidemic control~\cite{fox2022real, feng2020learning, zhang2020evaluating}, and intelligent transportation systems~\cite{feng2018deepmove, wang2024congestion, guo2024spatio, li2024st, jiang2025hcride, shao2022decoupled, SimpleTS}.
For instance, accurately forecasting bike-sharing demand enables service providers to optimize operations by preemptively redistributing bicycles to locations with anticipated high demand, thereby enhancing user satisfaction and system efficiency.

\begin{figure*}[ht]
  \centering  \includegraphics[width=\textwidth]{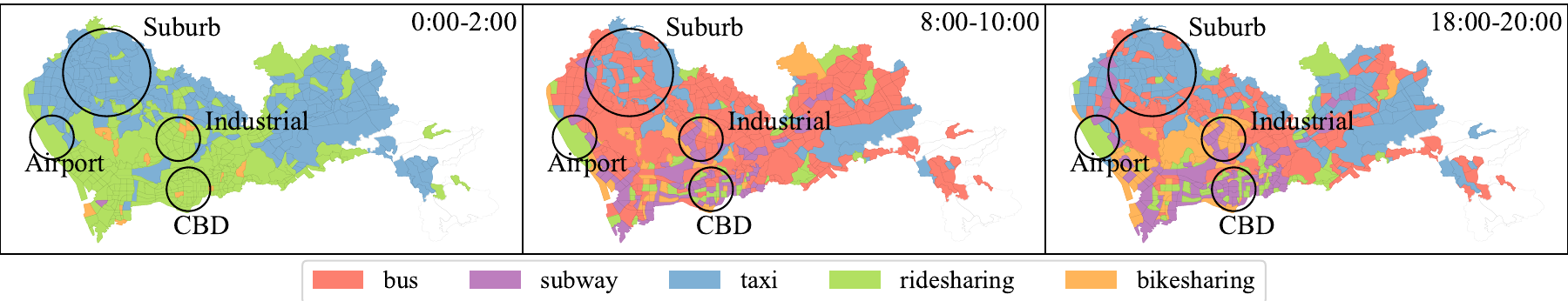}
  \caption{Dominant mobility mode in different regions at different hours in Shenzhen.}\vspace{-6pt}
  \label{Dominant Mode}
\end{figure*}

Due to its importance, numerous deep learning models have been proposed~\cite{stgcn,feng2018deepmove, zhou2023predicting, liang2022joint, zhao2023coupling} for spatiotemporal prediction in recent years.
However, most of these models are deterministic, which means they only focus on predicting expected mean values. These predictions may be unreliable and inaccurate because urban phenomena (e.g., various mobility modes, crime, and accidents) are highly dynamic and impacted by different contextual factors such as weather conditions, social events, and road closures. Although some recent works~\cite{stuanet,stzinb-gnn,uqunifi, jin2024spatial} propose probabilistic prediction models to quantify uncertainty for reliable prediction, most of them focus on a single phenomenon (e.g., taxi demand \cite{wang2020demand}, bikesharing mobility \cite{fang2021mdtp}, traffic flow \cite{han2024bigst}, crime \cite{li2022spatial}).
These models fail to capture inherent correlations of heterogeneous urban phenomena, which may potentially limit spatiotemporal prediction performance. 
For instance, during morning rush hours, a greater number of individuals in suburban areas might take buses from their homes to subway stations, subsequently taking subways to their workplaces. Conversely, in the evening rush hours, more people in industrial areas may opt to ride shared bikes to subway stations and then take subways back home.
Hence, capturing intrinsic spatiotemporal dynamics and complicated interactions between heterogeneous urban phenomena may lead to more accurate predictions. 

However, it is nontrivial to capture correlations of heterogeneous urban phenomena because they interact spatially and temporally and have inherent uncertainty. Also, it is impractical to assume that the data of different urban phenomena are independent and identically distributed (IID), which further makes the inferential statistics (such as hypothesis testing and confidence intervals) inappropriate for quantifying the uncertainty of multivariate spatiotemporal data.

To address these issues, we propose a novel uncertainty-aware Graph Neural Network framework called UQGNN for multivariate spatiotemporal prediction. In particular, UQGNN includes two innovative modules, i.e., an Interaction-aware Spatiotemporal Embedding module (ISTE) and a Multivariate Probabilistic Prediction module (MPP).
Within the ISTE module, we design a new Multivariate Diffusion Graph Convolutional Network (MDGCN) and an Interaction-aware Temporal Convolutional Network (ITCN) to capture the dynamics and dependencies of heterogeneous urban phenomena from spatial and temporal dimensions, respectively.
To deal with situations where samples are non-IID, we design an MPP module by introducing multivariate distributions to model interactions of heterogeneous urban phenomena.
The parameters in multivariate distributions define the characteristics of the distribution, such as location, spread, shape, and interrelationships between variables. For example, a multivariate Gaussian distribution is characterized by a mean vector representing expected values for each urban phenomenon and a covariance matrix indicating both the inherent uncertainty within each phenomenon and their interactions.
The key contributions of this paper are as follows:
\begin{itemize}
    \item \textbf{Conceptually}, this is the first study on uncertainty quantification of graph neural networks for multivariate spatiotemporal prediction, which aims to improve prediction performance by capturing complicated interactions and uncertainty of heterogeneous urban phenomena.
    
    \item \textbf{Technically}, we design a novel framework called UQGNN, which includes two key modules: (i) an ISTE module consisting of two innovative graph neural networks to model complex spatiotemporal interactions of heterogeneous urban phenomena, and (ii) an MPP module to quantify uncertainty and leverage it to further improve prediction accuracy.
    
    \item \textbf{Experimentally}, we extensively evaluate our UQGNN on four real-world multivariate spatiotemporal datasets from Shenzhen, New York City, and Chicago. We compare our UQGNN with 12 state-of-the-art baselines in terms of six metrics, and the experimental results demonstrate the superiority of our UQGNN, e.g., our UQGNN increases prediction accuracy by 5\% and uncertainty quantification by 5\% compared to the best baseline on the Shenzhen dataset. 
    The code is available at \blue{\url{https://github.com/UFOdestiny/UQGNN}}.

\end{itemize}

\section{Data Analysis and Motivation}\label{data}
In this section, we conduct a data-driven analysis on a multivariate spatiotemporal dataset including five different urban phenomena (mobility modes) in Shenzhen (i.e., taxi, bus, subway, bikesharing, and ridesharing mobility flow) to highlight key findings that motivate our work. Some findings are shown below.

\vspace{-10pt}
\begin{figure}[!htb]
\centering
\subfigure[CBD area]{
\label{fig:CBD}
\includegraphics[width=0.48\linewidth]{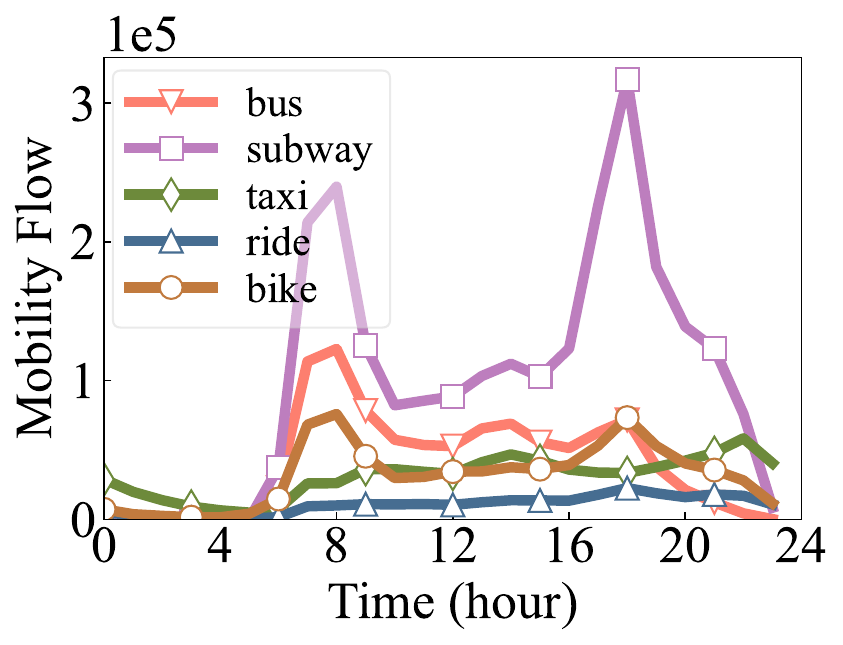}}
\subfigure[Industrial area]{
\label{fig:Industrial}
\includegraphics[width=0.48\linewidth]{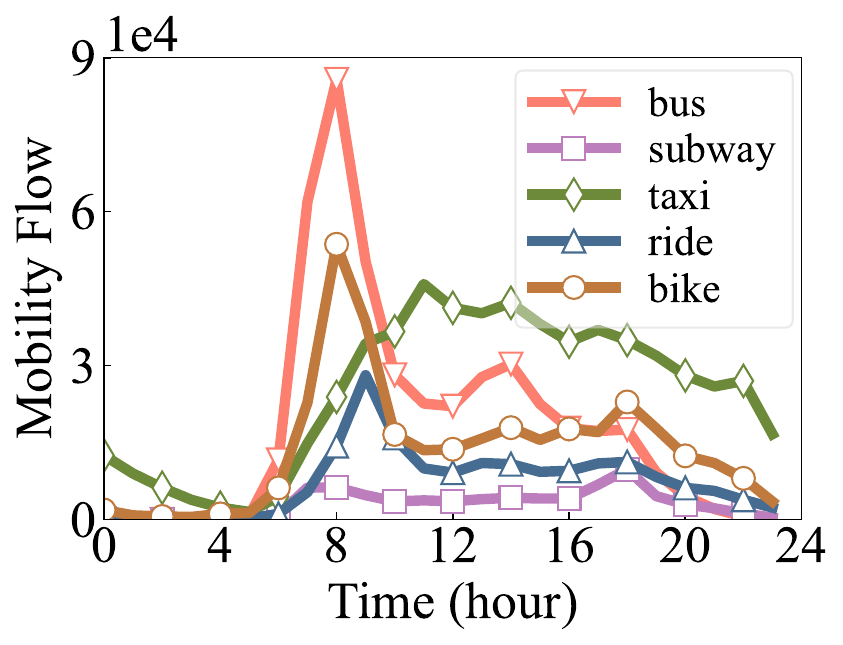}}
\vspace*{-6pt}
\caption{Five urban phenomena in different regions.}
\label{Inflow}
\vspace*{-8pt}
\end{figure}

(1) Dynamic Urban Phenomena. We found the dominant mobility mode in different regions usually changes at different hours. For example, as shown in Figure \ref{Dominant Mode}, the dominant mobility mode in the Central Business District (CBD) changed from the subway during rush hours to ridesharing at late night.
However, in some suburban areas, the bus and taxi are the dominant modes since there are no subway lines. During the morning rush hours, passengers in industrial areas tend to take buses, whereas bikesharing becomes more prevalent during the night rush hours. 
Additionally, some mobility modes exhibit a double-peak mobility phenomenon during 24 hours of a day (e.g., subway and bus in the CBD areas), contrasting with other modes like ridesharing in the industrial area that experience only a single peak, as shown in Figure~\ref{Inflow}.

\begin{figure}[!htb]
\vspace{-15pt}
\centering
\subfigure[PCC of CBD area]{
\includegraphics[width=0.47\linewidth]{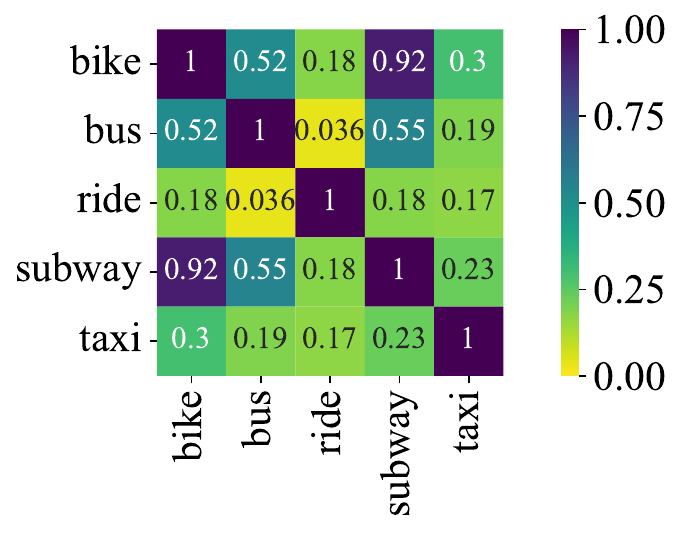}}
\hspace*{5pt}
\subfigure[PCC of industrial area]{
\includegraphics[width=0.47\linewidth]{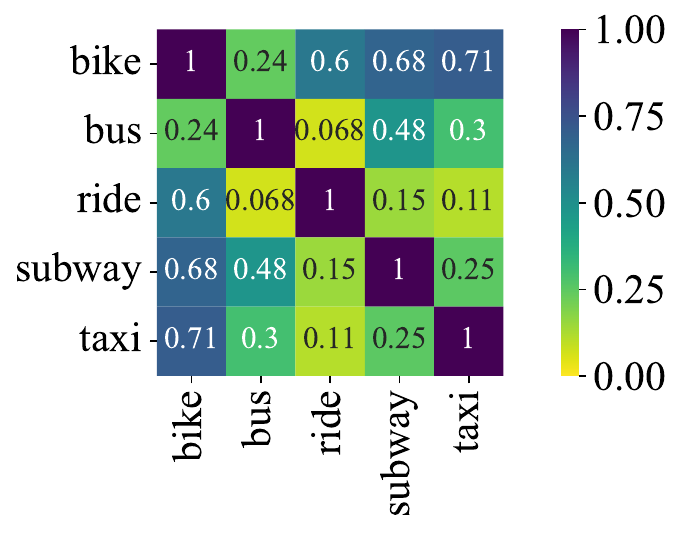}}
\vspace*{-13pt}
\caption{Correlations of five urban phenomena (mobility modes) at different regions.}
\label{pcc}
\vspace*{-10pt}
\end{figure}

(2) The correlations of urban phenomena vary in different regions. 
In Figure \ref{pcc}, we utilize the Pearson Correlation Coefficient (PCC) \cite{cohen2009pearson} to measure the correlations between different urban phenomena. 
Our analysis revealed that the PCC values of some phenomena in some regions are very large, for example, bikesharing and subway in the CBD area, which can potentially show their complementarity (e.g., riding a bike to a subway station).
Synchronously, the PCC values for certain modes, such as taxis and subways, are low, reflecting the infrequency with which people transfer from taxis to subways.

\textbf{Summary}: our data-driven investigation indicates the inherent dynamics of heterogeneous urban phenomena as well as their complicated correlations, which motivates us to capture those patterns for accurate multivariate spatiotemporal prediction.
In addition, existing works~\cite{he2023survey, stuanet} show that quantifying uncertainty is important for accurate and reliable spatiotemporal prediction. However, few existing studies focus on uncertainty quantification for heterogeneous urban data. To bridge this research gap, our work aims to enhance prediction accuracy and reliability by capturing the inherent interactions and uncertainties of different urban phenomena.

\section{Problem Formulation}\label{preliminary}
\subsection{Heterogeneous Urban Phenomena as A Graph}
A graph can be formally defined as $\mathcal{G}=\left( V, E, A \right)$, where $V$ is a set of nodes, $E$ is a set of edges, and $A$ is an adjacency matrix. In our setting, we consider each spatial region as a node, and each node has a set of features related to urban phenomena, e.g., mobility inflows, outflows, and the number of crimes in this region. Each edge represents the spatial distance between two nodes.
The adjacency matrix derived from the graph is denoted as $A\in \mathbb{R}^{N\times N}$, where $N$ is the number of nodes, and $A_{v_i,v_j}$ denotes the edge weight between region $v_i$ and region $v_j$. 
We adopt a Gaussian threshold kernel function to construct the adjacency matrix of our graph, which can be denoted as: 
\begin{equation}
A_{v_i,v_j}=\left\{\begin{array}{l} \exp(-\frac{d_{ij}^2}{\sigma^2}), i \neq j~\text{and}~\exp(-\frac{d_{ij}^2}{\sigma^2})>=r
\\ 0\quad\quad\quad\quad,\text{otherwise}.
\end{array}\right.
\end{equation}
where $d_{ij}$ represents the distance of the centroids between region $v_i$ and region $v_j$. $\sigma^2$ and $r$ are thresholds to control the distribution and sparsity of the matrix, where a large $r$ value can be used to accelerate the model training process. 
More details about the adjacency matrix will be shown in Appendix~\ref{adjacency}.

\subsection{Multivariate Spatiotemporal Prediction}
\subsubsection{Deterministic Prediction}

Given the above graph formulation, the goal of deterministic spatiotemporal prediction on graph $\mathcal{G}$ with $N$ nodes is to learn a mapping function $f$ from historical data of $t$ time steps $X_{1:t}=(x_{1},x_{2},...,x_{t})$ to predict values at future $T$ time steps, which can be expressed as follows:
\begin{equation}
X_{1:t}\stackrel{f}{\longrightarrow}\hat{X}_{t+1:t+T},
\end{equation}
where $\hat{X}_{t+1:t+T} \in \mathbb{R}^{N\times M\times T}$ and $M$ denotes the number of prediction variables (i.e., urban phenomena). 
Although existing research has greatly advanced deterministic prediction models for spatiotemporal prediction and achieved good prediction performance, it did not capture the prediction uncertainty caused by inherent correlations of multiple prediction variables (i.e., interactions of heterogeneous urban phenomena), which may potentially make unreliable predictions with low accuracy. 
Hence, probabilistic prediction models are necessary to deal with this challenge.

\subsubsection{Probabilistic Prediction}
Different from deterministic prediction models that only predict expected values, probabilistic prediction models leverage historical data $X_{1:t}$ as input to predict the distribution of future $T$ time steps, thus capturing the expectation and uncertainty of future urban phenomena. 
For example, if the input follows a multivariate Gaussian distribution, the output should also follow the same distribution, which indicates the predicted values and uncertainty. 
The process can be expressed as follows:
\begin{equation}
    \begin{aligned}
    &X_{1:t}\stackrel{g}{\longrightarrow}\hat{X}_{t+1:t+T},\\
    & \hat{X}_{t+1:t+T} \sim \mathcal N_M(\bm{\mu},\bm{\Sigma}),\\
    \end{aligned}
\end{equation}
where $X_{1:t} \in \mathbb{R}^{N\times M\times t}$. 
$M$ denotes the prediction variables (i.e., urban phenomena), $\bm{\mu}  \in \mathbb{R}^{N \times M \times T} $ and $\bm{\Sigma} \in \mathbb{R}^{N \times M \times M \times T}$ are used to quantify the predicted values and uncertainty, respectively.

\section{Methodology}\label{methodology}
\begin{figure*}[ht]
  \centering
  \includegraphics[width=\linewidth]{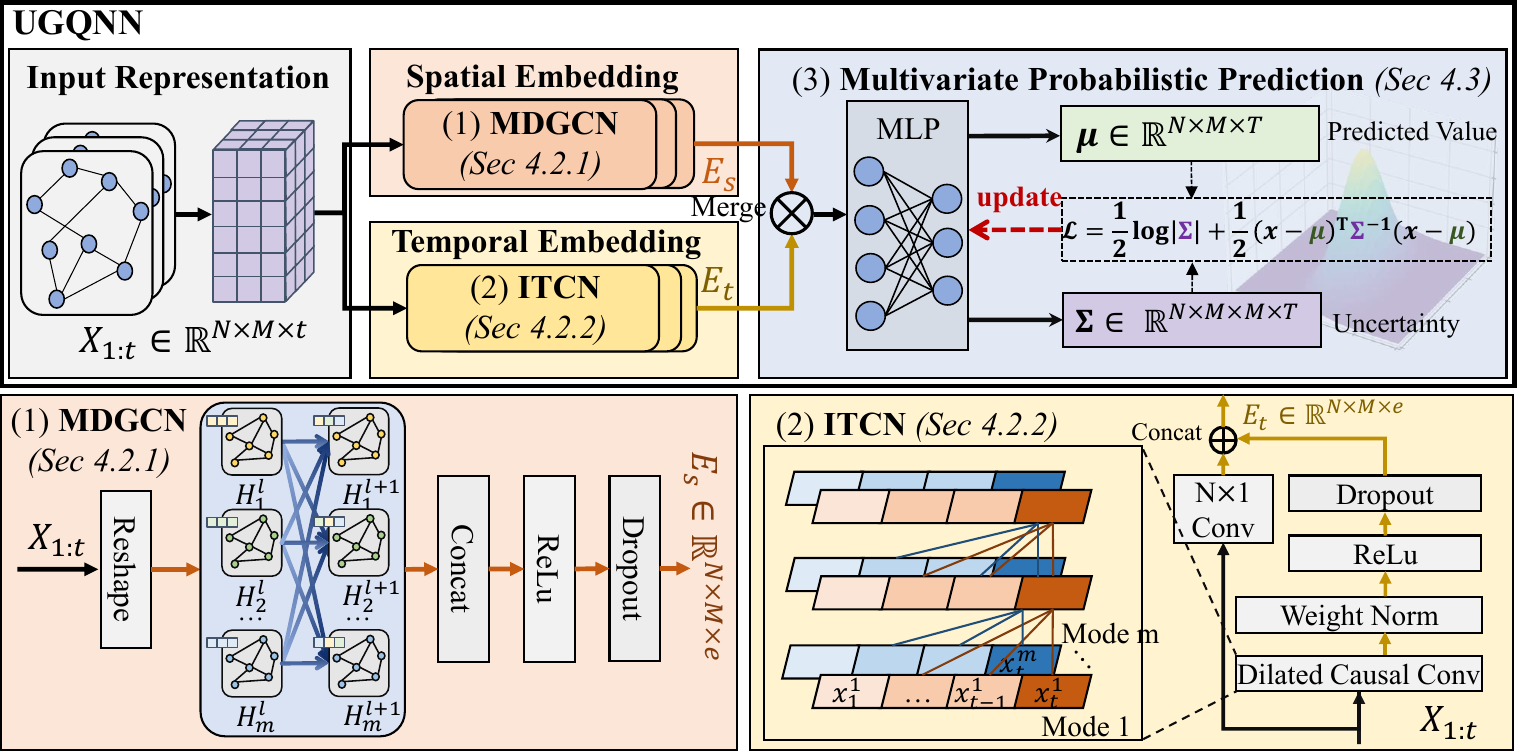}
  \caption{An overall framework of UQGNN. It consists of an Interaction-aware Spatiotemporal Embedding Module that includes (1) a spatial embedding component MDGCN, (2) a temporal embedding component ITCN, and (3) a multivariate probabilistic prediction (MPP) module. As for MDGCN, $H^l_m$ denotes the hidden state of the urban phenomenon $m$ in the $l$-th layer. The arrow between different states indicates an innovative cross-layer graph diffusion convolution, denoted in Equation~\ref{Eq.2}. The $x^m_t$ in the ITCN also indicates the urban phenomenon $m$ at time step $t$ as in Equation~\ref{Eq.3}.
   We can utilize different multivariate distributions to model the data based on data characteristics to perform probabilistic prediction and output parameters to represent mathematical quantities. For example, when we utilize the multivariate Gaussian distribution, the MPP layer will output $\bm{\mu}$ and $\bm{\Sigma}$, representing predicted values and uncertainty, respectively. The $\bm{\Sigma}$ is further used to update the parameters of the output layer by minimizing a well-designed negative log-likelihood loss function to improve the prediction accuracy. 
  }
  \label{model}
  \Description{}
  \vspace{-6pt}
\end{figure*}

\subsection{Overall Framework}
In this paper, we design a new probabilistic graph neural network framework called UQGNN for multivariate spatiotemporal prediction 
and also quantify uncertainty, as shown in Figure \ref{model}.

The original data is first processed to construct the graph, and the graph is then represented as a tensor $X_{1:t} \in \mathbb{R}^{N \times M \times t}$, where $N$ denotes the number of urban spatial regions, $M$ means the number of urban phenomena, and $t$ is the number of historical time steps.
After the input representation, we design two innovative modules for multivariate spatiotemporal prediction, i.e., (i) an \textbf{Interaction-aware Spatiotemporal Embedding (ISTE)} module and (ii) a \textbf{Multivariate Probabilistic Prediction (MPP)} module.
The ISTE module is designed to learn complicated spatiotemporal interactions of different urban phenomena and represent them as embeddings for prediction.
Two innovative GNNs are designed to capture spatial and temporal interaction patterns, respectively. 
The MPP module leverages the learned embeddings to predict probabilistic outputs (e.g., mean values and uncertainty), and more importantly, utilizes the predicted uncertainty to further improve the accuracy of the predicted values based on a well-designed loss function.  

\vspace{-3pt}
\subsection{Interaction-aware Spatiotemporal Embedding Module}
In this part, we design a Multivariate Diffusion Graph Convolutional Network and an Interaction-Aware Temporal Convolutional Network to capture spatial patterns and temporal patterns of heterogeneous urban phenomena, respectively. 

\subsubsection{Multivariate Diffusion Graph Convolutional Network}\label{mdgcn_model}

Motivated by that diffusion processes can effectively capture statistical dependencies between neighboring nodes in a network, in this work, we design a novel multivariate diffusion graph convolutional network (MDGCN) to learn spatial correlations from dynamic and heterogeneous inputs. This enables us to make a more comprehensive representation of how urban phenomena in one region influence those in nearby regions (e.g., transfer, complementarity, and competition), thereby improving prediction accuracy.

For most existing diffusion GCNs ~\cite{stzinb-gnn, dcrnn}, they consider samples at each node to be independent and identically distributed (IID).
However, in our scenario, various features (e.g., inflows of different mobility modes and traffic accidents) within a single node exhibit correlations, which means the IID assumption is no longer applicable, so relying on a single diffusion convolution is inadequate for modeling the correlations. To address this, we develop an MDGCN with cross-layer diffusion convolutions to capture interactions between different urban phenomena at each node.

We first model heterogeneous spatial dependencies of different urban phenomena with a diffusion process, which explicitly captures the stochastic nature of urban phenomenon dynamics~\cite{atwood2016diffusion, dcrnn}.
This diffusion process is characterized by a random walk on graph $\mathcal{G}$ with restart probability $\alpha \in [0,1]$ and a state transition matrix $D^{-1}_OA$, where $D_O = diag(A)$ is the out-degree diagonal matrix. 
After sufficiently large time steps, the Markov process 
(characterized by possible states and transition probabilities between those states) will converge to a stationary distribution $\mathcal{P} \in \mathbb{R}^{N\times N}$. 
Each row of $\mathcal{P}$ stands for a probability of diffusion from a node.
The stationary distribution of the diffusion process can be represented as a weighted combination of infinite random walks on the graph $\mathcal{G}$ with restart probability $\alpha \in [0,1]$, which can be calculated in the following closed form:
\begin{equation}
    \begin{aligned}
         \mathcal{P}=\sum_{k=0}^\infty \alpha(1-\alpha)^k(D_O^{-1}A)^k,
    \end{aligned}
    \label{Eq.1}
\end{equation}
where $k$ denotes the diffusion step, which is usually set to a finite number. According to Equation \ref{Eq.1}, we then define the backward diffusion process with backward transition matrix $\widetilde{W}_b=A^\intercal/\text{rowsum}(A^\intercal)$. Our adjacency matrix $A$ is symmetric, so $\widetilde{W}_f=\widetilde{W}_b$. 
The forward and backward diffusion processes can model the dynamics of urban phenomena. For example, the diffusion processes that model between residential areas and CBD areas can be interpreted as commuting in the morning and evening. 

As shown in our data-driven analysis in Section~\ref{data}, there are complicated and varying correlations between different urban phenomena, so it is not suitable to utilize graph diffusion for each mode separately and add them together. To address this, we 
design a cross-layer diffusion convolution by applying graph diffusion to all urban phenomena in other layers.
The resulting cross-layer diffusion convolution operation of MDGCN over a graph signal and a filter is defined as follows: 
\begin{equation}
    \begin{aligned}
    &H^{l+1}_m=\sigma\sum_{m=1}^M\left(\sum_{k=1}^K (T_k(\widetilde{W}_f)H^l_m\Theta^{k}_{f,l}+T_k(\widetilde{W}_b)H^l_m\Theta^{k}_{b,l})\right),\\
    \end{aligned}
    \label{Eq.2}
\end{equation}
where $m$ denotes the number of urban phenomenon types, $H^l$ represents the $l^{th}$ hidden layer; learned parameters $\Theta^k_{f,l}$ and $\Theta^k_{b,l}$ in the $l^{th}$ layer are added to control how each node transforms the received information; $\sigma$ is an activation function (e.g., ReLU or Linear); The Chebyshev polynomial is used to approximate the convolution operation since it is challenging to obtain directly, which can be represented as $T_k(X)=2AT_{k-1}(X)-T_{k-2}(X)$, with boundary conditions $T_0(X)=I, T_1(X)=X$.
For each layer, the convolution operation is applied across all the dimensions.
In this way, each layer will fully leverage the information from other urban phenomena to capture comprehensive spatial correlations. 
We stack multiple MDGCN layers to better capture spatial dependencies of different urban phenomena and output the spatial embedding $E_s \in \mathbb{R}^{N \times M \times e}$, where $e$ denotes the dimension of spatial embeddings.

\subsubsection{Interaction-Aware Temporal Convolutional Network}
Temporal Convolutional Network (TCN)~\cite{lea2017temporal, lea2016temporal} has been widely adopted for temporal modeling due to its capability of efficiently capturing long-term dependencies and flexible receptive field. 
However, conventional TCNs operate under the assumption of channel independence, ignoring the mutual correlations among different channels, so they cannot be directly used for modeling interactions of heterogeneous urban phenomena. Therefore, 
in this paper, we design a new Interaction-aware Temporal Convolutional Network (ITCN), specifically engineered to harness information from other dimensions to effectively model correlations. The dilated convolution operation $F$ on element $s$ of the sequence can be expressed as: 
\begin{equation}
    \begin{aligned}
F(s) = \sum^{M}_{m=1}\sum^{k-1}_{i=0} f(i) \cdot X_{s-d\times i}^{m},
    \end{aligned}\label{Eq.3}
\end{equation}
where $X$ is the input sequence, $f$ is a convolutional filter, $d$ is the dilation factor, $k$ is the filter size, and $s-d\times i$ accounts for the direction of the past.
Therefore, each ITCN layer $H_l$ will receive the signals from the previous layer, which is updated using:
\begin{equation}
    \begin{aligned}
    H_l=f(\Gamma_l *H_{l-1}+b),
    \end{aligned}
\end{equation}
where $\Gamma_l$ is the convolution filter for the corresponding layer, $*$ is the shared convolution operation, and $b$ stands for the bias.
In this way, each type of urban phenomenon will fully leverage the information from other types of urban phenomena. 
In our model, we stack multiple ITCN layers to capture the temporal dynamics and heterogeneity better and output the temporal embedding $H_T$. 

As shown in Figure~\ref{model}, to make our ITCN more than just an overly complex linear regression model, activation functions are added on top of the convolutional layers to introduce non-linearities. ReLU activations are added to the residual blocks after both convolutional layers.
Weight normalization is applied to each convolutional layer to normalize the input of hidden layers (which counteracts the exploding gradient problem, among other benefits).
In order to prevent overfitting, regularization is introduced via dropout after each convolutional layer in every residual block. 
We also stack multiple ITCN layers, and this module will finally output the temporal embeddings represented as $E_t \in \mathbb{R}^{N \times M \times e}$, where $e$ denotes the dimension of temporal embeddings.

Finally, we utilize the Hadamard product to integrate spatial and temporal embeddings as $(E) _{ij}=(E_s  \otimes E_t) _{ij}=(E_s) _{ij}(E_t) _{ij}$ since it emphasizes both spatially and temporally important features, which is widely adopted by prior works \cite{dstagnn,stuanet} and was proven as an effective method to fuse spatial and temporal embeddings.

\subsection{Multivariate Probabilistic Prediction}
In this paper, one of our key contributions is to improve prediction accuracy by quantifying the inherent uncertainty of heterogeneous urban phenomena, which involves utilizing the learned uncertainty to help make more accurate predictions. 
Motivated by our data analysis, we consider different multivariate distributions to describe the data of heterogeneous urban phenomena, so we can leverage a parameter in the distribution (e.g., covariance in multivariate Gaussian distribution) to quantify the uncertainties of different urban phenomena and their interactions.
To be more specific, suppose we utilize a multivariate Gaussian distribution to model the node input variable $X$ to capture the interactions between different types of urban phenomena, in which the mean vector $\bm{\mu}$ provides expected values for each urban phenomenon. The diagonal entries of the covariance matrix $\bm{\Sigma}$ capture uncertainties within individual urban phenomena and uncertainties caused by their interactions.

\begin{algorithm}[!ht]
\caption{Multivariate Probabilistic Prediction}
\begin{algorithmic}[1]
\REQUIRE Fused spatiotemporal embedding $E$; minimum eigenvalue threshold $V_{min}$; output blocks $H_\mu$ and $H_z$.
\ENSURE Mean vector $\mu$; positive definite matrix $\Sigma$.
\FOR{each timestep $\tau$ in $[t+1, t+T]$}
\FOR{each region $n$ in $[1, N]$}
\STATE Compute mean vector $\mu_{\tau;n}\in \mathbb{R}^{M} \leftarrow H_\mu(E_n)$
\STATE Compute vector $Z_{\tau;n}\in \mathbb{R}^{(M+1)\times M/2} \leftarrow H_\Sigma(E_n)$
\STATE Initialize empty matrix $W\in \mathbb{R}^{M\times M}$.
\STATE Copy $Z$ to $W$ by the indices of the upper triangle.
\STATE Copy $Z$ to $W$ by the indices of the lower triangle.
\STATE Compute eigenvalues $E_{val}$ and eigenvectors $E_{vec}$ of $Z$.
\STATE Clamp eigenvalues $E_{val}$ with the minimum value of $V_{min}$ to ensure positivity and get $E_{val}'$.
\STATE Construct intermediate matrix $I=E_{vec}\times E_{val}'$.
\STATE Reconstruct positive definite matrix $\Sigma_{\tau;n}=I\times E_{vec}^\intercal$.
\ENDFOR
\ENDFOR
\RETURN $\mu_{t+1:t+T;1:N}$, $\Sigma_{t+1:t+T;1:N}$
\end{algorithmic}
\label{alg}
\vspace{-4pt}
\end{algorithm}

As shown in Algorithm \ref{alg}, we first define the output block consisting of a temporal convolutional layer, two linear layers, a normalization layer, and a ReLU activation layer. For each region and each time step, two output blocks $H_\mu$ and $H_z$ with different dimensions are designed to compute the $\mu$ and $Z$ separately.
It is important to note that the covariance matrix must be symmetric. Thus, we only need to output a vector of dimensions ${N\times (N+1)/2}$, representing half of the matrix, including the diagonal. Additionally, the covariance matrix must be positive definite, meaning all its eigenvalues should be positive. However, directly learning such a matrix through neural networks is challenging.
To address this, we initialize an empty matrix $W\in \mathbb{R}^{M\times M}$ and copy $Z$ to $W$ by the indices of the upper and lower triangles, thereby ensuring the matrix is symmetric.
Next, we compute the eigenvalues $E_{val}$ and the eigenvectors $E_{vec}$ of $Z$. The eigenvalues $E_{val}'$ are clamped according to the pre-defined minimum eigenvalue threshold $V_{min}$, which is typically set between $10^{-6}$ and $10^{-2}$.
If $V_{min}$ is set too small, the eigenvalues of the matrix may be over-restricted, which will limit the expressiveness of the model and affect the training effect. This may cause eigenvalues close to zero or negative eigenvalues in the covariance matrix, which will cause the failure of subsequent calculations (e.g., inverse matrix and eigen-decomposition). While $V_{min}$ is set too large, large eigenvalues may cause the model to ignore details in the data and limit its degrees of freedom, affecting the training effect and accuracy of the model.
Finally, we construct an intermediate matrix $I=E_{vec}\times E_{val}'$ and reconstruct the positive definite matrix $\Sigma_{t;n}=I\times E_{vec}^\intercal$.

Taking the multivariate Gaussian distribution as an example, we assume that the probability of ground truth $X$ given a model input $\hat{X}$ can be approximated by a multivariate Gaussian distribution:
\begin{equation}
    \begin{aligned}        p(X;\bm{\mu},\bm{\Sigma})=\frac{\exp[-\frac{1}{2}(\hat{X}-\bm{\mu})^\intercal\bm{\Sigma}^{-1}(\hat{X}-\bm{\mu})]}{\sqrt{(2\pi)^D|\bm{\Sigma}|}}
    \end{aligned}
\end{equation}
where $\bm{\mu}$ is the mean of $\hat{X}$ and $\bm{\Sigma}$ is the covariance matrix. However, directly maximizing the predictive Gaussian likelihood is numerically unstable because multiplying small values together can lead to arithmetic underflow, the situation in which the digital representation of a floating point number reaches its limit.
Instead, we try to maximize the following log-likelihood. In the training process, we leverage the negative logarithm of the likelihood to minimize the loss function of the result as follows:
\begin{equation}
    \begin{aligned}
    \mathcal{L}(X;\bm{\mu},\bm{\Sigma})=\frac{1}{2}\log|\bm{\Sigma}|+\frac{1}{2}(X-\bm{\mu})^\intercal\bm{\Sigma}^{-1}(X-\bm{\mu})
    \end{aligned}
\end{equation}

The negative log-likelihood (NLL) loss function penalizes deviations between the model's predictions and the actual data of the Gaussian distribution using the Adam optimization algorithm~\cite{zhang2018improved}. 
The update rule involves subtracting a fraction of the gradients from the current parameter values, scaled by a learning rate hyperparameter. 
By iteratively updating the parameters of the multivariate Gaussian distribution to minimize the NLL loss, the model learns to represent the underlying correlation of the data better.

We leverage the multivariate Gaussian distribution to model the correlation between various urban phenomena rather than regions or time steps because we are more concerned about the correlation between different urban phenomena after spatiotemporal representation, and we assume the heterogeneous urban phenomena follow a multivariate Gaussian distribution. 

More importantly, UQGNN is not only applicable to the multivariate Gaussian distribution but also to other multivariate distributions, such as the multivariate Laplace and multivariate negative binomial distributions. More details are shown in Appendix~\ref{algo}.




\section{Evaluation}\label{evaluation}
In this section, we conduct a comprehensive experimental evaluation of our proposed framework. Specifically, we aim to address the following five research questions:
\begin{itemize}[leftmargin=5.0mm]
\item \textbf{RQ 1}: Is our UQGNN more effective than other baselines?
\item \textbf{RQ 2}: Is our UQGNN effective for uncertainty quantification?
\item \textbf{RQ 3}: Are all components in UQGNN framework effective?
\item \textbf{RQ 4}: Is it more effective for prediction by capturing interactions of different urban phenomena (i.e., variables)?
\item \textbf{RQ 5}: How do various distributions impact model performance?
\end{itemize}
\vspace{-6pt}

\begin{figure*}[!htb]
\centering
\subfigure[DiffSTG]{
\label{pres1}
\includegraphics[width=0.24\textwidth]{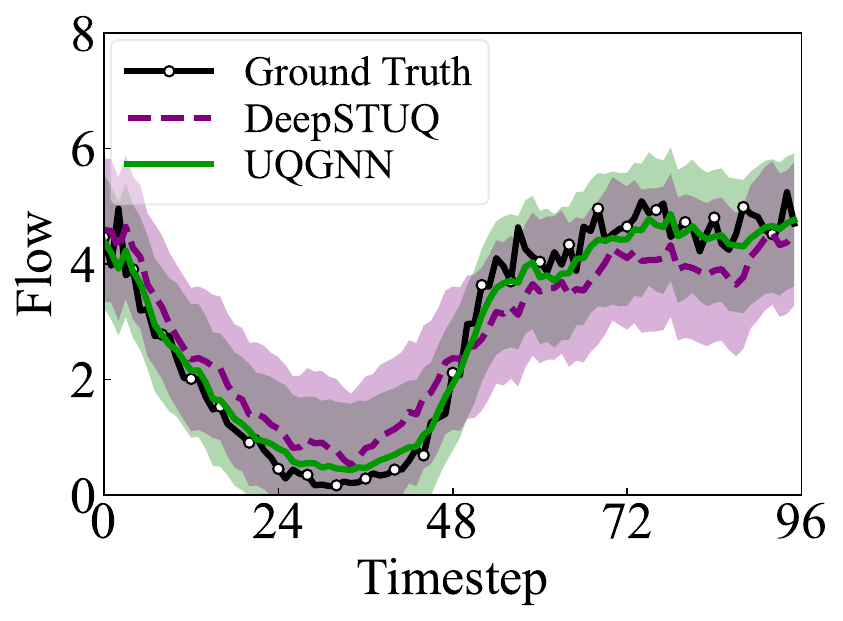}}
\subfigure[STZINB]{
\label{pres2}
\includegraphics[width=0.24\textwidth]{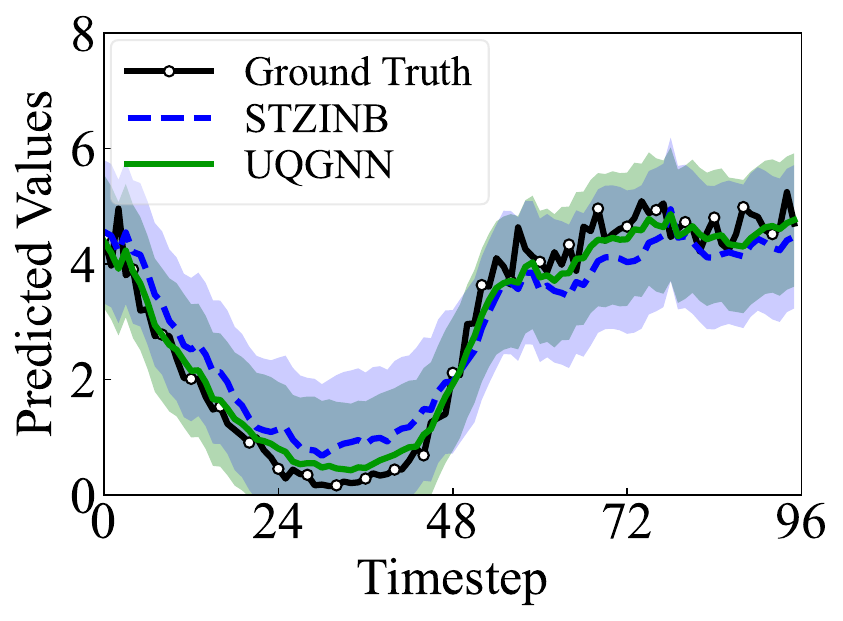}}
\subfigure[CF-FNN]{
\label{pres4}
\includegraphics[width=0.24\textwidth]{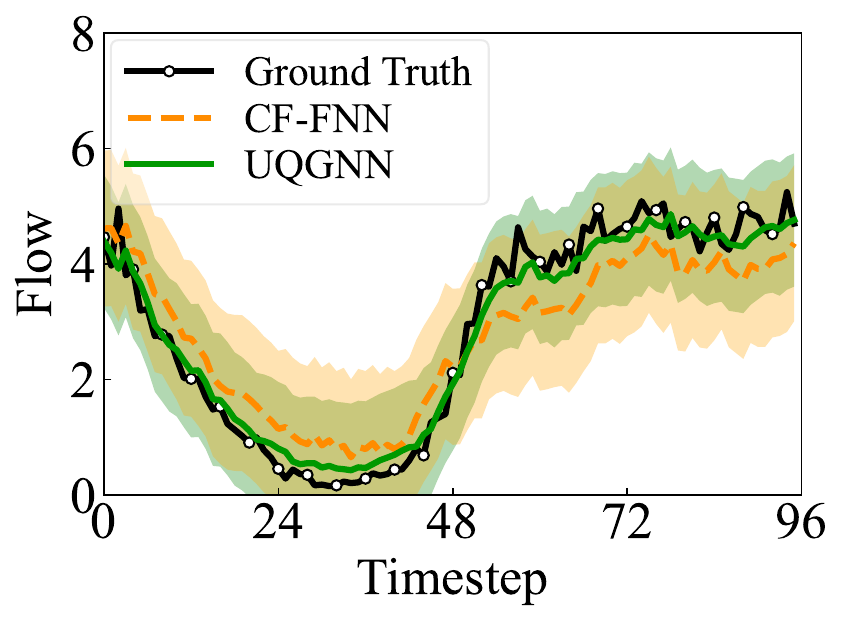}}
\subfigure[DeepSTUQ]{
\label{pres5}
\includegraphics[width=0.24\textwidth]{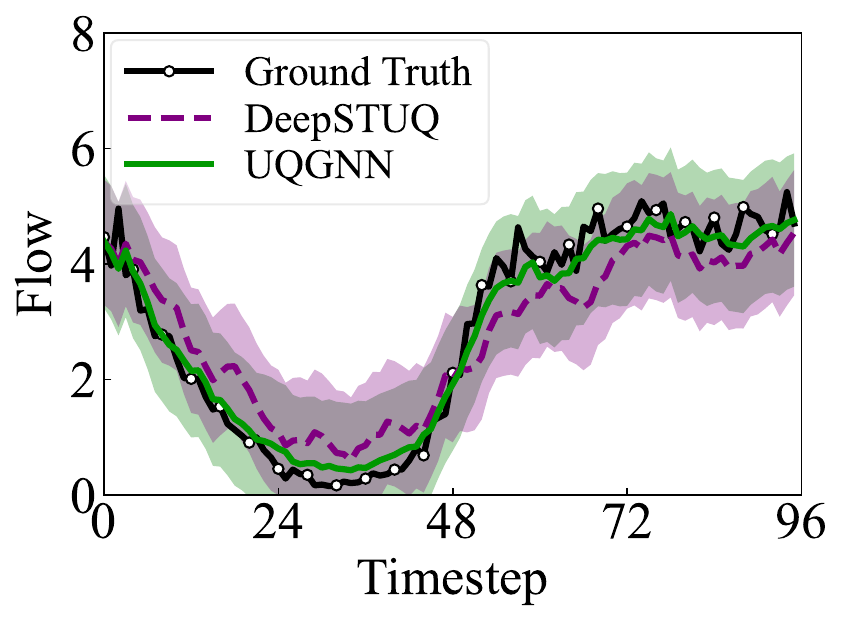}}
\vspace*{-8pt}
\caption{Prediction results of the taxi flow on the NYC Dataset 1 in terms of the last 96 timesteps, i.e., 24 hours, with best baselines selected for visualization. The shadow represents the results of probabilistic prediction in terms of the MPIW.}
\label{probres}
\vspace{-10pt}
\end{figure*}

\vspace{-3pt}
\subsection{Experiment Setup}
\subsubsection{Datasets}
We evaluate UQGNN on four real-world heterogeneous urban datasets from Shenzhen, New York City (NYC), and Chicago.
The Shenzhen dataset includes five mobility modes (bus, subway, taxi, ridesharing, bikesharing) across 491 regions, recorded hourly from January to June 2019.
For NYC, the first dataset covers taxi~\cite{nyc_taxi}, bikesharing~\cite{nyc_bike}, and subway~\cite{nyc_subway}; the second includes crime~\cite{nyc_crime} and crash~\cite{nyc_crash} data, both recorded every 15 minutes in 67 Manhattan zones from January to June 2022. The Chicago dataset (taxi~\cite{chi_taxi}, bikesharing~\cite{chi_bike}, and crime~\cite{chi_crime}) spans 77 regions with 15-minute records from September to November 2022.

\subsubsection{Baselines}
We compare our UQGNN with 12 state-of-the-art baselines, including 7 deterministic spatiotemporal prediction models (i.e., 
Spatial-Temporal Graph Convolution Network (STGCN)~\cite{stgcn}, 
Diffusion Convolutional Recurrent Neural Net (DCRNN)~\cite{dcrnn},
Graph WaveNet (GWNET)~\cite{GraphWavenet}, 
Spectral Temporal Graph Neural Network (StemGNN)~\cite{cao2020spectral},
Dynamic Spatial-Temporal Aware Graph Neural Network (DSTAGNN)~\cite{dstagnn},  
Adaptive Graph Convolutional Recurrent Network (AGCRN)~\cite{bai2020adaptive}, and 
SUMformer~\cite{cheng2024rethinking}), 
and 5 probabilistic spatiotemporal prediction models (i.e., 
TimeGrad~\cite{rasul2021autoregressive},
Spatial-Temporal Zero-Inflated Negative Binomial Graph Neural Network (STZINB)~\cite{stzinb-gnn}, 
Deep Spatio-Temporal Uncertainty Quantification (DeepSTUQ)~\cite{uqunifi},
DiffSTG~\cite{wen2023diffstg}, and 
CF-CNN~\cite{huang2024uncertainty}). 
Detailed descriptions of these baselines are provided in Appendix \ref{baseli}.



\begin{table*}[]
\setlength{\tabcolsep}{12pt}
\centering 
\caption{Comparison with 12 state-of-the-art baselines on four real-world datasets in terms of six metrics. The best results are presented in bold, and the second-best
results are underlined.
}\vspace{-5pt}
\begin{tabular}{l|l|cccccc}
\hline
\multirow{2}{*}{Datasets} & \multirow{2}{*}{Methods} & \multicolumn{6}{c}{Metrics} \\ \cline{3-8} 
 &  & MAE & RMSE & MAPE & KL & MPIW & CRPS \\ \hline
\multirow{13}{*}{Shenzhen Dataset} 
 & STGCN~\cite{stgcn} & 11.831 & 55.055 & 0.532 & 7.255 & 31.403 & 15.848 \\
  & DCRNN~\cite{dcrnn} & 11.554 & 42.437 & 0.602 & 7.265 & 33.251 & 13.291 \\
& GWNET~\cite{GraphWavenet} & 12.681 & 59.612 & 0.526 & 9.259 & 47.136 & 14.352 \\
 & StemGNN~\cite{cao2020spectral} & 10.485 & 42.301 & 0.602 & 7.032 & 32.442 & 11.755 \\
 & DSTAGNN~\cite{dstagnn} & 10.272 & 36.524 & 0.539 & 7.153 & 33.523 & 11.756 \\
 & AGCRN~\cite{bai2020adaptive} & 12.527 & 37.125 & \underline{0.523} & 9.286 & 33.751 & 16.836 \\
 & SUMformer~\cite{cheng2024rethinking} & 11.742 & 38.293 & 0.594 & 7.356 & \underline{31.263} & 13.177 \\
 & TimeGrad~\cite{rasul2021autoregressive} & 11.580 & 54.275 & 0.546 & 8.136 & 35.427 & 12.543 \\
  & STZINB~\cite{stzinb-gnn} & 10.343 & 35.376 & 0.587 & 8.213 & 31.281 & \underline{9.562} \\
 & DiffSTG~\cite{wen2023diffstg} & \underline{10.053} & \underline{34.591} & 0.581 & \underline{6.848} & 33.832 & 10.421 \\
 & DeepSTUQ~\cite{uqunifi} & 11.522 & 37.318 & 0.650 & 8.021 & 42.291 & 14.267 \\
 & CF-GNN~\cite{huang2024uncertainty} & 10.371 & 38.183 & 0.591 & 7.575 & 32.749 & 9.913 \\
 & \textbf{UQGNN} & \textbf{9.717} & \textbf{32.273} & \textbf{0.503} & \textbf{6.555} & \textbf{29.342} & \textbf{8.399} \\ \hline
\multirow{13}{*}{NYC Dataset 1} 
& STGCN~\cite{stgcn} & 6.194 & 10.966 & 0.305 & 3.834 & 35.021 & 6.119 \\
 & DCRNN~\cite{dcrnn} & 6.243 & 11.042 & 0.324 & \underline {3.208} & 36.274 & 6.175 \\
& GWNET~\cite{GraphWavenet} & 6.396 & 11.804 & 0.312 & 3.424 & 38.262 & 6.214 \\
  & StemGNN~\cite{cao2020spectral} & 6.252 & 11.356 & 0.315 & 3.355 & 37.539 & \underline {5.974} \\
 & DSTAGNN~\cite{dstagnn} & 6.315 & 11.461 & 0.310 & 3.375 & 35.740 & 6.013 \\
 & AGCRN~\cite{bai2020adaptive} & 6.571 & 12.088 & 0.305 & 3.372 & 37.152 & 6.227 \\
 & SUMformer~\cite{cheng2024rethinking} & 6.232 & 10.957 & 0.314 & 3.553 & 34.137 & 6.102 \\
 & TimeGrad~\cite{rasul2021autoregressive} & \underline{6.184} & 12.647 & 0.333 & 3.922 & 35.212 & 6.323 \\
 & STZINB~\cite{stzinb-gnn} & 6.248 & 11.261 & 0.305 & 3.455 & 34.107 & 6.181 \\
 & DiffSTG~\cite{wen2023diffstg} & 6.223 & \underline{10.842} & \underline {0.293} & 3.831 & 35.403 & 6.234 \\
 & DeepSTUQ~\cite{uqunifi} & 6.193 & 11.318 & 0.327 & 3.311 & 35.143 & 6.153 \\
 & CF-GNN~\cite{huang2024uncertainty} & 6.201 & 12.364 & 0.312 & 3.272 & \underline{33.728} & 6.001 \\

 & \textbf{UQGNN} & \textbf{5.982} & \textbf{10.775} & \textbf{0.278} & \textbf{3.053} & \textbf{33.711} & \textbf{5.883} \\ \hline
\multirow{13}{*}{NYC Dataset 2} 
 & STGCN~\cite{stgcn} & 6.044 & 10.935 & 0.299 & 3.718 & 34.170 & 5.978 \\
 & DCRNN~\cite{dcrnn} & 6.061 & 11.153 & 0.318 & 3.122 & 35.548 & 5.815 \\

& GWNET~\cite{GraphWavenet} & 6.021 & 11.128 & 0.307 & \underline{3.030} & 37.174 & \underline{5.794} \\
 & StemGNN~\cite{cao2020spectral} & 6.089 & 11.035 & 0.308 & 3.067 & 36.413 & 5.814 \\
 & DSTAGNN~\cite{dstagnn} & 6.134 & 11.032 & 0.304 & 3.100 & 34.914 & 5.892 \\
  & AGCRN~\cite{bai2020adaptive} & 6.349 & 11.846 & 0.299 & 3.092 & 36.037 & 5.798 \\
 & SUMformer~\cite{cheng2024rethinking} & \underline{5.966} & \underline{10.667} & 0.308 & 3.069 & 33.164 & 5.940 \\

 & TimeGrad~\cite{rasul2021autoregressive} & 6.000 & 12.248 & 0.326 & 3.804 & 34.155 & 6.197 \\
 & STZINB~\cite{stzinb-gnn} & 6.085 & 10.923 & 0.299 & 3.351 & 33.145 & 6.057 \\
 & DiffSTG~\cite{wen2023diffstg} & 5.980 & 10.828 & \underline{0.287} & 3.716 & 34.341 & 6.049 \\
 & DeepSTUQ~\cite{uqunifi} & 6.003 & 11.015 & 0.320 & 3.211 & 34.090 & 6.028 \\
 & CF-GNN~\cite{huang2024uncertainty} & 6.062 & 12.016 & 0.307 & 3.756 & \textbf{32.935} & 5.881 \\

 & \textbf{UQGNN} & \textbf{5.843} & \textbf{10.486} & \textbf{0.272} & \textbf{2.980} & \underline{32.937} & \textbf{5.707} \\ \hline
\multirow{13}{*}{Chicago Dataset} 
 & STGCN~\cite{stgcn} & 1.752 & 3.116 & 0.615 & 1.840 & 2.905 & \textbf{0.664} \\
  & DCRNN~\cite{dcrnn} & 1.741 & 3.614 & \underline{0.597} & 1.371 & 2.451 & 0.720 \\
& GWNET~\cite{GraphWavenet} & 1.828 & 3.623 & 0.713 & 1.923 & 2.857 & 0.945 \\
 & StemGNN~\cite{cao2020spectral} & 1.704 & 3.710 & 0.624 & 1.368 & 2.687 & 0.756 \\
 & DSTAGNN~\cite{dstagnn} & 1.747 & 3.437 & 0.654 & 1.268 & 2.374 & 0.685 \\
 & AGCRN~\cite{bai2020adaptive} & 1.742 & 3.213 & 0.640 & 1.354 & 2.576 & 0.835 \\
 & SUMformer~\cite{cheng2024rethinking} & \textbf{1.672} & 3.403 & 0.616 & 1.358 & 2.812 & 0.722 \\
 & TimeGrad~\cite{rasul2021autoregressive} & 1.705 & 3.588 & 0.665 & \underline{1.213} & 2.604 & 0.683 \\
 & STZINB~\cite{stzinb-gnn} & 1.685 & 3.103 & 0.707 & 1.235 & 2.513 & 0.727 \\
 & DiffSTG~\cite{wen2023diffstg} & 1.807 & 3.279 & 0.625 & 1.231 & 2.464 & 0.696 \\
 & DeepSTUQ~\cite{uqunifi} & \underline{1.673} & \underline{3.024} & 0.608 & 1.267 & 2.431 & 0.715 \\
 & CF-GNN~\cite{huang2024uncertainty} & 1.685 & 3.114 & 0.613 & 1.242 & \textbf{2.324} & 0.732 \\

 & \textbf{UQGNN} & 1.680 & \textbf{3.012} & \textbf{0.590} & \textbf{1.189} & \underline{2.333} & \underline{0.680} \\ \hline

\end{tabular}\label{restable}
\end{table*}

\vspace{-3pt}
\subsubsection{Evaluation Metrics}

We leverage three widely used \textbf{\textit{deterministic metrics}}, including Mean Absolute Error (MAE), Root Mean Squared Error (RMSE), and Mean Absolute Percentage Error (MAPE), to evaluate the performance of deterministic prediction. 
In addition, since traditional metrics for accuracy are not directly applicable to probabilistic prediction, we also utilize three other commonly used \textbf{\textit{probabilistic metrics}}, including Continuous Ranked Probability Score (CRPS), Kullback-Leibler Divergence (KL), and Mean Prediction Interval Width (MPIW), to evaluate the uncertainty quantification performance of different models. Detailed descriptions are shown in Appendix~\ref{allmetrics}

\vspace{-8pt}
\subsection{Overall Performance Comparison (RQ 1)}
A comprehensive comparison of our UQGNN and other baseline models is presented in Table \ref{restable}. We found UQGNN consistently achieves the best performance across almost all metrics on the four datasets. Specifically, UQGNN improves MAE by approximately 5\% on Shenzhen dataset, 3\% on both NYC Dataset 1 and NYC Dataset 2, and 2\% on Chicago dataset compared to the best baseline.
For probabilistic prediction, UQGNN also demonstrates superior accuracy, with a 5\% improvement in CRPS on Shenzhen dataset, 3\% on the NYC dataset 1, 3\% on the NYC dataset 2, and 2\% on the Chicago dataset. 
The improved accuracy and reduced error demonstrate that UQGNN is not only effective in prediction but also capable of capturing nuanced patterns in complex multivariate data.

As shown in Figure \ref{probres}, UQGNN’s deterministic predictions align most closely with the ground truth. In the probabilistic setting, the shadow around the curve represents the prediction interval, where a tighter shadow indicates more reliable predictions. It is observed that the green shadow of our UQGNN is more compact while still covering most observations, indicating more reliable and precise estimates compared to baselines.



\vspace{-12pt}
\subsection{Effectiveness of UQ (RQ 2)}
Furthermore, we leverage selective regression~\cite{sokol2024conformalized, shah2022selective} to further explain the results and demonstrate the uncertainty quantification performance.
Selective regression allows abstention from prediction if the confidence is not sufficient. Two quantities characterize the performance of selective regression:
(i) \emph{coverage}, i.e., the fraction of samples that the model makes predictions on. For example, if the model rejects the top 10\% of samples because of their higher uncertainty score, the coverage will be 90\%; and (ii) \emph{error}, which is represented by MAE in this work.

\begin{figure}[!htb]
\centering
\subfigure[Shenzhen dataset]{
\includegraphics[width=0.48\linewidth]{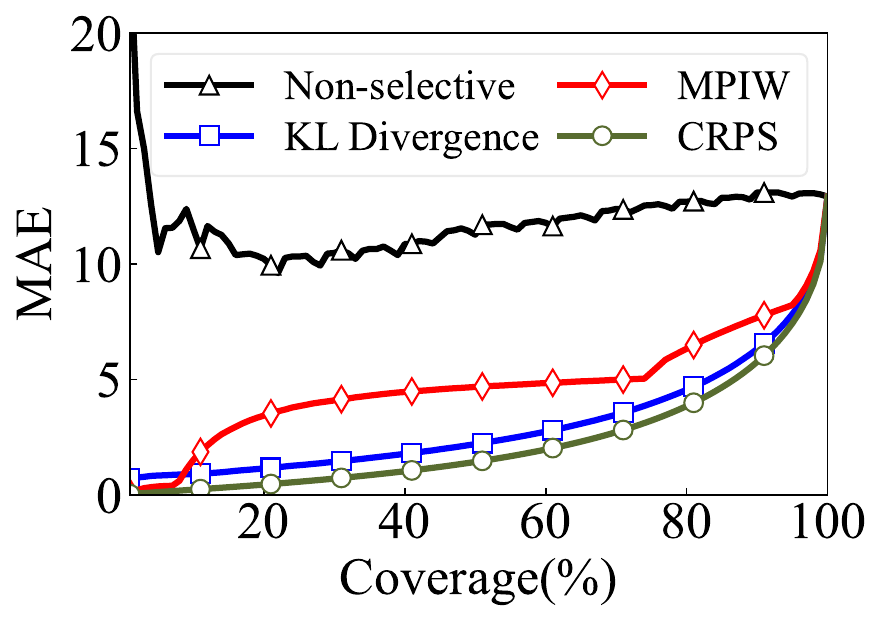}}\vspace{-8pt}
\subfigure[NYC dataset 1]{
\includegraphics[width=0.48\linewidth]{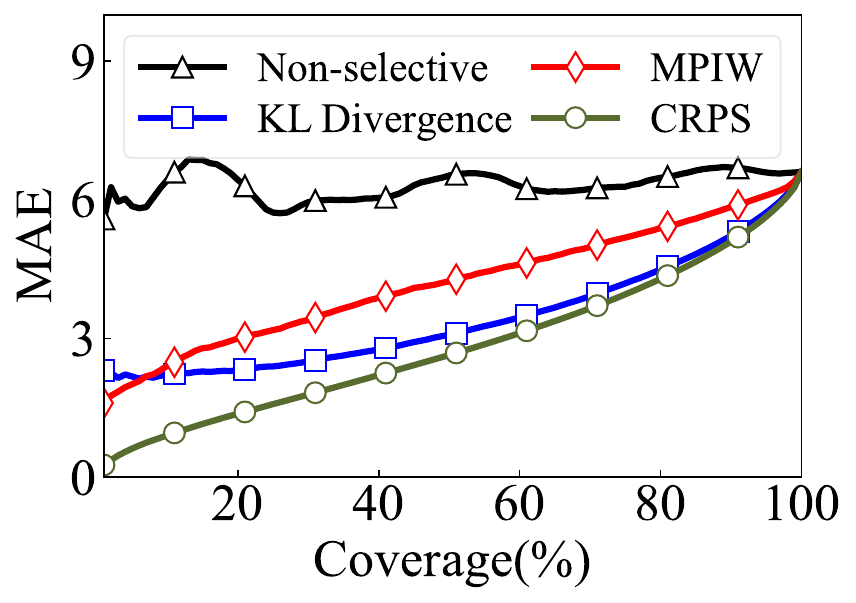}}
\subfigure[NYC dataset 2]{
\includegraphics[width=0.48\linewidth]{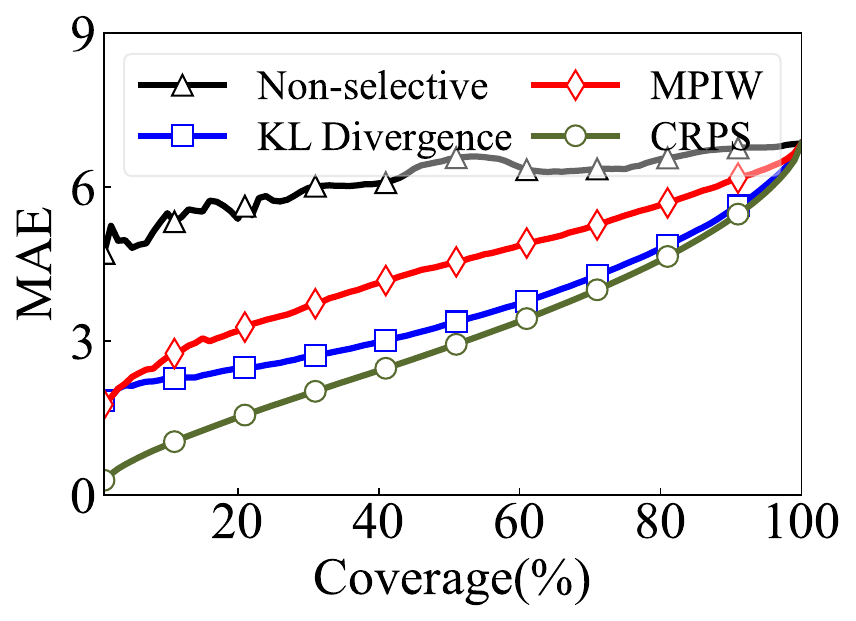}}
\subfigure[Chicago dataset]{
\includegraphics[width=0.48\linewidth]{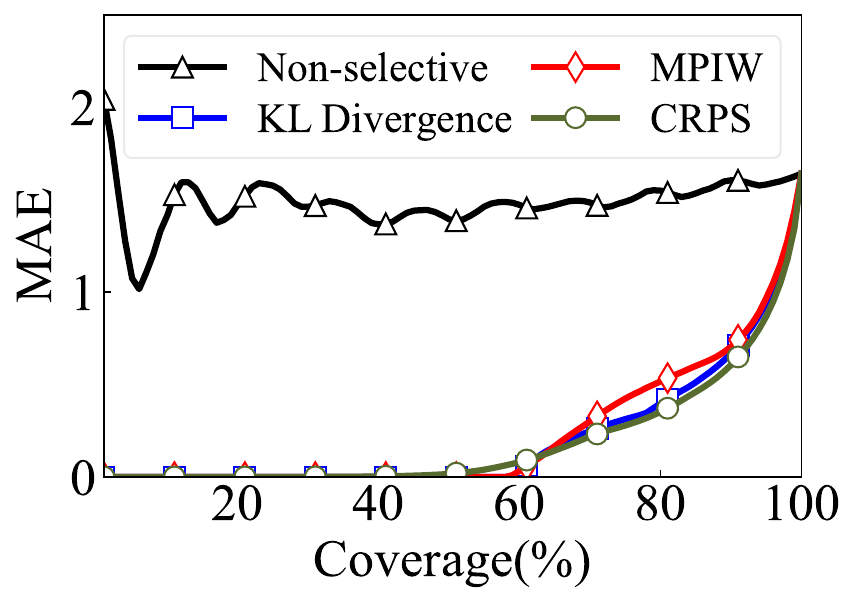}}
\vspace{-8pt}
\caption{Selective regression for our UQGNN.}
\label{selective}
\vspace*{-14pt}
\end{figure}

As shown in Figure \ref{selective}, MAE remains almost horizontal without considering uncertainty quantification for all the datasets, indicating that the prediction error is nearly irrelevant to the coverage.
However, when uncertainty is considered, the error increases along with coverage. 
This indicates the prediction error is positively correlated with the coverage, so the uncertainty we calculated is meaningful and instructive. Thus, the effectiveness of the uncertainty quantification of our UQGNN is verified.

\vspace{-6pt}

\vspace{-6pt}
\subsection{Ablation Study (RQ 3)}
We conduct an ablation study to show the effectiveness of each key component in UQGNN. Specifically, we compare UQGNN with four variants by removing a component in UQGNN. The four variants are explained below:
\begin{itemize}[leftmargin=5.0mm]
\item \textbf{w/o MDGCN}: by replacing the MDGCN with traditional diffusion GCN, which means without capturing the spatial interactions of heterogeneous urban phenomena.
\item \textbf{w/o ITCN}: by replacing the ITCN with traditional TCN, which means without capturing the temporal interactions of different urban phenomena.
\item \textbf{w/o MPP}: by removing the multivariate probabilistic prediction, which means only capturing the spatiotemporal interactions without considering uncertainty.
\item \textbf{w/ MPP-}: by replacing the multivariate Gaussian distribution with multiple univariate Gaussian distributions to describe various urban phenomena.
\end{itemize}

From Table \ref{ablation1}, we can see that the performance of the model without the MDGCN significantly decreases (e.g., more than 20\% decrease in MAE on the Shenzhen dataset), which indicates the importance of capturing the spatial interactions of different urban phenomena. 
Similarly, the performance degradation of UQGNN w/o ITCN also indicates the effectiveness of our ITCN design and the importance of capturing temporal interactions.
In addition, the performance decreased without the MPP module, which indicates our uncertainty quantification method can help effectively improve prediction accuracy. Furthermore, when we replace the multivariate Gaussian distribution with multiple univariate Gaussian distributions, the performance also decreases since it cannot capture the inherent uncertainties caused by interactions.

\vspace{-6pt}
\begin{table}[htbp] \small
\setlength{\tabcolsep}{2pt}
\caption{Ablation study of different design components.} 
\centering
\vspace{-6pt}
\begin{tabular}{l|c|cccccc}
\hline
\multirow{2}*{Data} & \multirow{2}{*}{Methods} & \multicolumn{5}{c}{Metrics} \\\cline{3-8}
		~ &  & MAE & RMSE& MAPE& KL & MPIW &CRPS \\
\hline
 & w/o MDGCN & 13.717 & 38.422 & 0.825 & 9.575 & 31.292 & 11.52 \\
 & w/o ITCN & 18.572 & 41.510 & 0.723 & 8.923 & 37.423 & 15.47 \\
Shenzhen & w/o MPP & 13.213 & 38.359 & 0.684 & 8.051 & 31.654 & 10.48 \\
 & w/ MPP- & 12.341 & 38.578 & 0.682 & 7.521 & 34.310 & 10.58 \\
 & \textbf{UQGNN} & \textbf{9.717} & \textbf{32.273} & \textbf{0.503} & \textbf{6.555} & \textbf{29.342} & \textbf{8.399} \\ \hline
 & w/o MDGCN & 11.657 & 16.139 & 0.378 & 3.491 & 43.460 & 8.761 \\
 & w/o ITCN & 7.932 & 15.761 & 0.365 & 3.273 & 39.631 & 7.962 \\
NYC 1 & w/o MPP & 6.885 & 14.986 & 0.343 & 3.563 & 37.413 & 6.378 \\
 & w/ MPP- & 7.862 & 11.551 & 0.312 & 3.724 & 38.252 & 6.554 \\
 & \textbf{UQGNN} & \textbf{5.982} & \textbf{10.775} & \textbf{0.278} & \textbf{3.053} & \textbf{33.711} & \textbf{5.883} \\ \hline
 & w/o MDGCN & 11.025 & 16.232 & 0.421 & 3.827 & 52.989 & 9.642 \\
 & w/o ITCN & 7.542 & 14.467 & 0.457 & 3.659 & 41.710 & 5.919 \\
NYC 2 & w/o MPP & 6.738 & 13.513 & 0.435 & 3.192 & 35.277 & 6.356 \\
 & w/ MPP- & 7.954 & 12.051 & 0.423 & 3.325 & 38.457 & 6.681 \\
 & \textbf{UQGNN} & \textbf{5.843} & \textbf{10.486} & \textbf{0.272} & \textbf{2.980} & \textbf{32.937} & \textbf{5.707} \\ \hline
 & w/o MDGCN & 4.792 & 7.113 & 0.812 & 1.846 & 8.261 & 0.778 \\
 & w/o ITCN & 2.547 & 4.924 & 0.697 & 1.574 & 6.737 & 0.843 \\
Chicago & w/o MPP & 2.373 & 4.531 & 0.688 & 1.358 & 5.714 & 0.807 \\
 & w/ MPP- & 2.030 & 4.215 & 0.845 & 1.283 & 4.493 & 0.751 \\
 & \textbf{UQGNN} & \textbf{1.680} & \textbf{3.012} & \textbf{0.590} & \textbf{1.189} & \textbf{2.333} & \textbf{0.680} \\ \hline
\end{tabular}
\label{ablation1}
\vspace{-5pt}
\end{table}

\vspace{-5pt}
\subsection{Importance of Capturing Interactions (RQ 4)}
We also compare the performance of UQGNN for the prediction of heterogeneous urban phenomena with that of individual urban phenomenon prediction. 
The results in Table \ref{ablation3} show that considering heterogeneous urban phenomena can significantly improve the prediction performance of all urban phenomena compared to predicting any individual urban phenomenon. 

\begin{figure}[!htb]
\centering
\subfigure[Shenzhen Dataset]{
\includegraphics[width=0.48\linewidth]{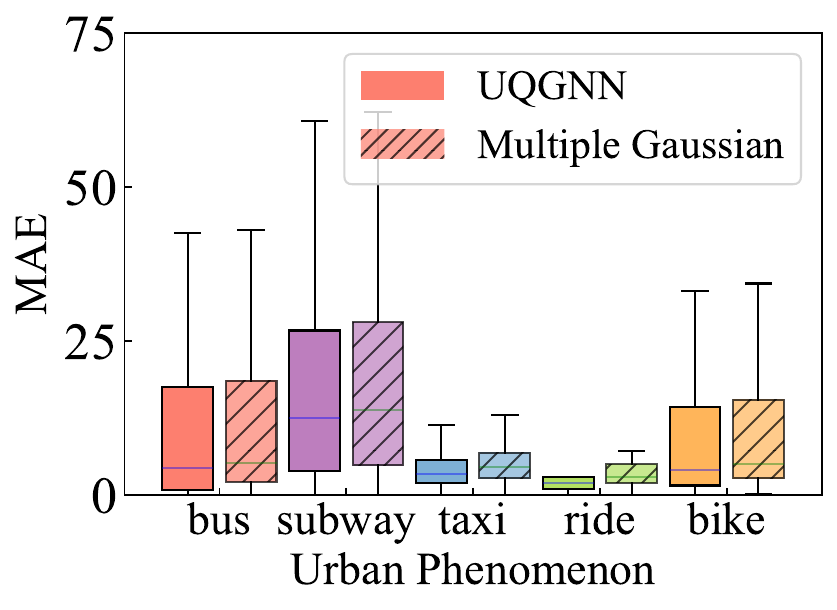}}\vspace{-5pt}
\subfigure[NYC Dataset 1]{
\includegraphics[width=0.48\linewidth]{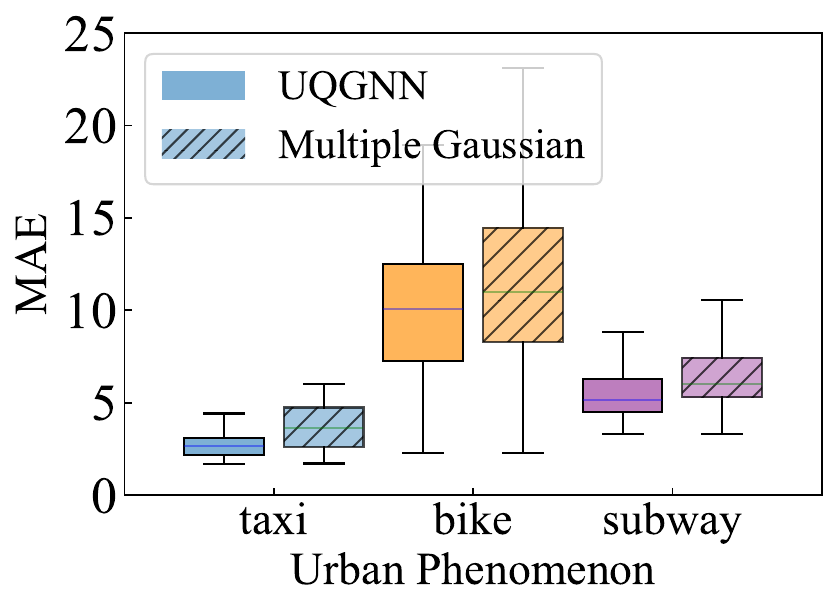}}
\subfigure[NYC Dataset 2]{
\includegraphics[width=0.48\linewidth]{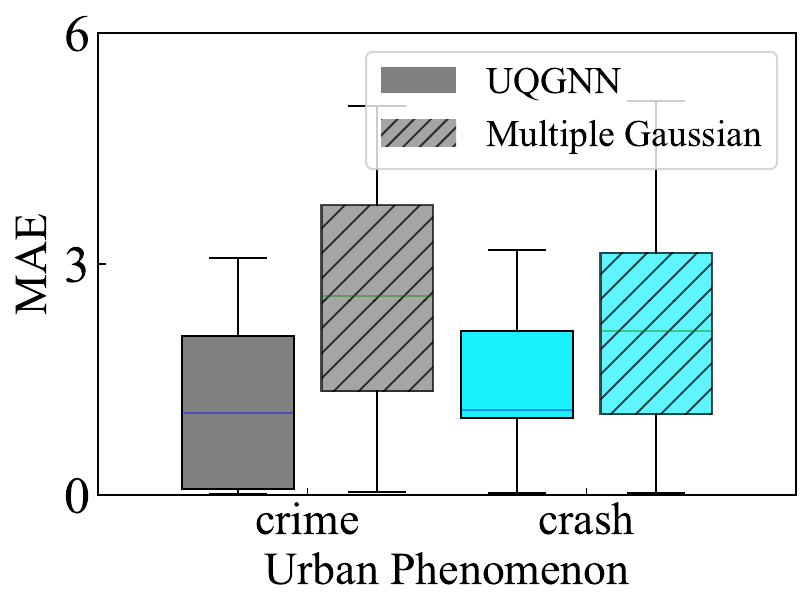}}
\subfigure[Chicago Dataset]{
\includegraphics[width=0.48\linewidth]{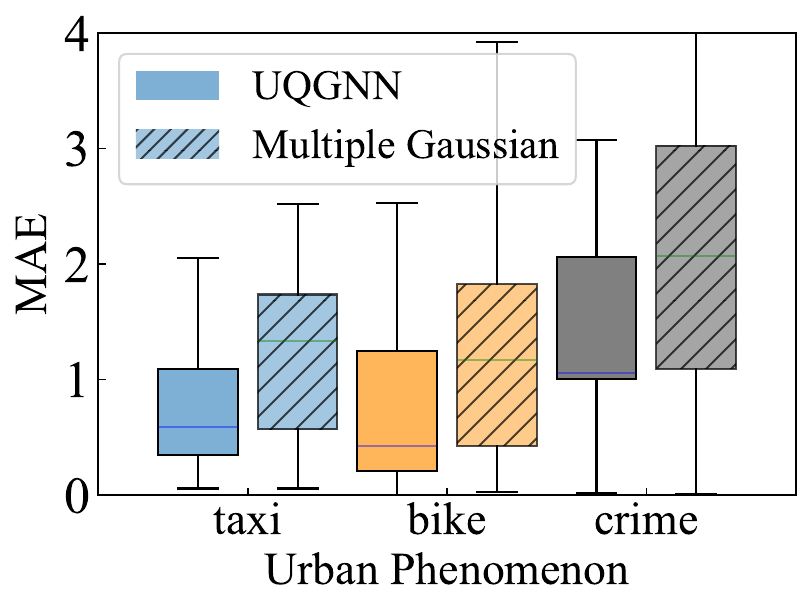}}
 \vspace*{-10pt}
\caption{Prediction error of each urban phenomenon.}
\label{Prediction Error m}
\vspace{-15pt}
\end{figure}

In addition, Figure \ref{Prediction Error m} shows the percentage increase in MAE of UQGNN compared to predicting individual urban phenomena separately. The MAE of each urban phenomenon prediction with UQGNN using a multivariate Gaussian distribution is lower than predicting them separately. This also demonstrates that considering heterogeneous urban phenomena can improve the prediction performance of all the urban phenomena compared to predicting any individual phenomenon since more interaction patterns and uncertainty can be captured.
\begin{table}[htbp]\small
\setlength{\tabcolsep}{2.5pt}
\caption{Improvement for each urban phenomenon.}\vspace{-5pt}
\centering

\begin{tabular}{l|c|cccccc}
\hline
\multirow{2}*{Datasets} & \multirow{2}{*}{Phenomena} & \multicolumn{5}{c}{Increase on Each Metric} \\\cline{3-8}
		~ &  & MAE & RMSE & MAPE  & KL & MPIW & CRPS\\
\hline
&bus        &6.1\% & 16.5\% & 15.0\% & 7.0\% & 25.8\% & 37.2\% \\
&subway &3.0\% & 9.8\%  & 14.3\% & 6.2\% & 17.7\% & 24.4\% \\
Shenzhen&taxi   &7.2\% & 17.0\% & 28.8\% & 9.4\% & 30.4\% & 44.8\% \\
&ridesharing&5.1\% & 10.7\% & 12.7\% & 5.6\% & 20.0\% & 28.4\% \\
&bikesharing&5.9\% & 11.9\% & 12.4\% & 4.8\% & 25.0\% & 34.1\% \\

\hline
\multirow{3}{*}{NYC 1} &taxi    &2.9\% & 9.2\%  & 14.0\% & 2.1\% & 9.1\%  & 13.8\% \\
&bikesharing&8.5\% & 17.6\% & 23.8\% & 6.0\% & 20.6\% & 38.4\% \\
&subway    &5.2\% & 9.4\%  & 13.3\% & 2.3\% & 9.1\%  & 15.7\% \\

\hline
\multirow{2}{*}{NYC 2} 
&crime&1.6\% & 15.1\% & 25.2\% & 5.1\% & 21.3\% & 39.9\% \\
&crash&1.7\% & 14.7\% & 28.3\% & 4.6\% & 17.8\% & 38.7\% \\
\hline
&taxi    &2.4\% & 8.1\%  & 13.1\% & 2.1\% & 10.6\% & 14.6\% \\
Chicago&bikesharing&7.7\% & 18.1\% & 26.6\% & 4.7\% & 22.2\% & 37.8\% \\
&crime&7.4\% & 14.6\% & 25.7\% & 4.8\% & 21.3\% & 40.6\% \\

\bottomrule
\end{tabular}

\vspace{-15pt}
\label{ablation3}
\vspace{-0pt}
\end{table}

\vspace{-3pt}

\subsection{Impact of Distributions (RQ 5)}\label{appendix_distribution}
We further study the impact of different distributions in the MPP module. Particularly, we compare five different distributions:
\begin{itemize}[leftmargin=5.0mm]
\item \textbf{Laplace}: using multivariate Laplace distributions in the probabilistic prediction module. 
\item \textbf{T-distribution}: using multivariate Student's \emph{t} distributions. T-distribution is symmetric around zero and bell-shaped.
\item \textbf{Negative Binomial}: using multivariate negative binomial distributions, which model the number of failures in a sequence of IID Bernoulli trials.
\item \textbf{Poisson}: using the multivariate Poisson distribution, which models the number of events within a specific period.
\item \textbf{Gaussian}: using multivariate Gaussian distribution to model relationships among multiple variables.
\end{itemize}

As shown in Table~\ref{ablation2}, we found that using the multivariate Gaussian distribution achieves the best performance on all four datasets compared with other distributions. In particular, it improves the MAE, RMSE, and MAPE by around 10\% and also increases the CRPS, KL, and MPIW by 7\%. This is reasonable since the data distributions of many urban phenomena follow Gaussian distributions, e.g., mobility flows as shown in Figure~\ref{Inflow}. Our UQGNN framework is inherently generalizable and can be adapted to accommodate any distribution, while still allowing for the consideration of specific data characteristics to optimize performance.


\vspace{-5pt}
\begin{table}[htbp] \small
\setlength{\tabcolsep}{1.5pt}
\caption{Comparison of different multivariate distributions.} \vspace{-5pt}
\centering
\begin{tabular}{l|c|cccccc}
\hline
\multirow{2}*{Datasets} & \multirow{2}{*}{Distributions} & \multicolumn{5}{c}{Metrics} \\\cline{3-8}
		~ &  & MAE & RMSE & MAPE & KL & MPIW& CRPS \\
\hline
 & Laplace & 16.216 & 42.053 & 0.783 & 7.637 & 36.056 & 14.292 \\
 & T-distribution & 17.132 & 43.226 & 0.717 & 7.688 & 41.222 & 15.449 \\
Shenzhen & Negative Binomial & 14.687 & 38.255 & 0.685 & 7.715 & 35.350 & 16.278 \\
 & Poisson & 13.680 & 37.342 & 0.612 & 7.483 & 34.521 & 10.601 \\
 & \textbf{Gaussian} & \textbf{9.717} & \textbf{32.273} & \textbf{0.503} & \textbf{6.555} & \textbf{29.342} & \textbf{8.399} \\ \hline
 & Laplace & 8.452 & 12.908 & 0.381 & 4.282 & 41.586 & 6.914 \\
 & T-distribution & 8.917 & 13.340 & 0.401 & 4.179 & 42.476 & 7.378 \\
NYC 1 & Negative Binomial & 8.503 & 12.922 & 0.352 & 4.081 & 40.615 & 7.286 \\
 & Poisson & 6.532 & 10.786 & 0.363 & 3.709 & 37.261 & 6.815 \\
 & \textbf{Gaussian} & \textbf{5.982} & \textbf{10.775} & \textbf{0.278} & \textbf{3.053} & \textbf{33.711} & \textbf{5.883} \\ \hline
 & Laplace & 8.533 & 12.877 & 0.423 & 3.512 & 40.014 & 6.912 \\
 & T-distribution & 8.855 & 13.585 & 0.445 & 3.813 & 39.953 & 7.421 \\
NYC 2 & Negative Binomial & 8.487 & 11.925 & 0.401 & 3.637 & 38.174 & 6.028 \\
 & Poisson & 6.892 & 10.643 & 0.352 & 3.105 & 36.482 & 6.332 \\
 & \textbf{Gaussian} & \textbf{5.843} & \textbf{10.486} & \textbf{0.272} & \textbf{2.980} & \textbf{32.937} & \textbf{5.707} \\ \hline
 & Laplace & 2.077 & 4.131 & 0.689 & 0.318 & 3.117 & 0.696 \\
 & T-distribution & 3.418 & 5.206 & 1.203 & 0.857 & 3.335 & 0.872 \\
Chicago & Negative Binomial & 2.525 & 4.674 & 0.917 & 0.634 & 3.931 & 0.983 \\
 & Poisson & 2.163 & 5.013 & 0.742 & 0.359 & 2.667 & 0.781 \\
 & \textbf{Gaussian} & \textbf{1.680} & \textbf{3.012} & \textbf{0.590} & \textbf{1.189} & \textbf{2.333} & \textbf{0.680} \\ \hline
\end{tabular}
\vspace{-6pt}
\label{ablation2}
\end{table}

\section{Related Work}\label{literature review}
Spatiotemporal prediction is important for many real-world applications such as urban planning~\cite{jiang2018deepurbanmomentum,chen2024enhancing}, transportation optimization~\cite{feng2018deepmove, liu2023st4ml, yuan2022route, liu2020multi}, pandemic control~\cite{fox2022real, luo2020deeptrack}, and emergency response~\cite {song2014prediction}. 
Recently, many studies have been conducted to improve spatiotemporal prediction driven by the advances of deep learning techniques. 
We summarize related works into four categories from two dimensions: 
homogeneous vs. heterogeneous; 
uncertainty-agnostic vs. uncertainty-aware, as shown in Table \ref{ResearchGap}.

\vspace{-5pt}
\begin{table}[h]
\caption{A taxonomy of spatiotemporal prediction.} 
\vspace{-5pt}
\begin{tabular}{l|c|c}
\hline
 & Homogeneous & Heterogeneous \\ 
\hline
Uncertainty-agnostic &~\cite{wang2019origin,stgcn,feng2018deepmove, zhou2023predicting,dstagnn}&~\cite{ liu2021co, liang2022joint, zhao2023coupling, qiu2024tfb} \\ \hline
Uncertainty-aware &~\cite{stuanet,stzinb-gnn,uqunifi, jin2024spatial, tudm, sauc}  & \textbf{UQGNN}\\ 
\hline
\end{tabular}
\label{ResearchGap}
\vspace{-11pt}
\end{table}

\subsection{Uncertainty-agnostic Spatiotemporal Prediction}

Most existing spatiotemporal prediction works only focus on deterministic prediction of a single urban phenomenon (e.g., taxi or crash).
For example, Lin \textit{et al.}~\cite{lin2020self} propose a novel self-attention memory to memorize features with long-range dependencies of spatial and temporal domains for taxi traffic flow prediction.
Graph Convolutional Recurrent Network (GCRN)~\cite{GCRN} has been proposed for spatiotemporal prediction, aiming to concurrently recognize spatial structures and dynamic variations in structured sequences. The primary challenge is to identify the most effective combinations of recurrent networks and graph convolution within specific settings.
Yu \textit{et al.}~\cite{stgcn} design a novel deep learning framework called STGCN for traffic prediction based on loop detector data, which integrates graph convolution and gated temporal convolution through spatiotemporal convolutional blocks.
DCRNN~\cite{dcrnn} leverages the diffusion process to model the spatial correlation characterized by a random walk on the given graph with a probability.

In recent years, some works have also focused on multivariate spatiotemporal prediction~\cite{ye2019co, liu2021co, liang2022joint, zhao2023coupling, kang2022rawlsgcn, zhao2023multiple}. 
For instance, Liu \textit{et al.}~\cite{liu2021co} design a self-learned spatial graph construction to predict both taxi and bikesharing mobility together, and Ye \textit{et al.}~\cite{ye2019co} predict both taxi and bikesharing mobility with deep CNN and heterogeneous LSTM.
Ding \textit{et al.} \cite{deng2024multi} design a self-supervised learning approach to predict both bike and taxi demand. 
However, few of these works capture prediction uncertainty, which is not only significant for prediction-based decision-making but also has the potential to help improve prediction accuracy.

\vspace{-2pt}
\subsection{Uncertainty-aware Spatiotemporal Prediction}

In recent years, uncertainty quantification has attracted much interest from the spatiotemporal prediction community because it can lay the foundation for reliable and safe decision-making, which is important for practical applications like transportation optimization and crime prevention~\cite{stzinb-gnn, jiang2025uncertainty, huang2024uncertainty}. 
Wen \textit{et al.}~\cite{wen2023diffstg} generalize the popular denoising diffusion probabilistic models to spatiotemporal graphs called DiffSTG to capture the intrinsic uncertainties.
Kashif \textit{et al.}~\cite{rasul2021autoregressive} propose TimeGrad, an autoregressive model for multivariate probabilistic time series forecasting which samples from the data distribution at each time step by estimating its gradient.
Kexin \textit{et al.}~\cite{huang2024uncertainty} extend conformal prediction to graph-based models for guaranteed uncertainty estimates.
Zhuang \textit{et al.}~\cite{stzinb-gnn} design a Spatial-Temporal Zero-Inflated Negative Binomial Graph Neural Network called STZINB to quantify the uncertainty of sparse urban data.
Qian \textit{et al.}~\cite{uqunifi} develop DeepSTUQ to quantify the uncertainty of traffic prediction with two independent sub-neural networks.


Nevertheless, most existing works on uncertainty-aware spatiotemporal prediction only consider a single urban phenomenon without capturing interactions between different urban phenomena. Our data-driven results show both opportunities and challenges to utilize data from one urban phenomenon to enhance the prediction of another phenomenon caused by their complicated interactions, which motivates us to design new uncertainty-aware GNNs for multivariate spatiotemporal prediction.

\section{Conclusion}\label{conclusion}
In this paper, we propose an uncertainty-aware graph neural network framework called UQGNN for multivariate spatiotemporal prediction. There are two key novel designs in UQGNN, i.e., an Interaction-aware Spatiotemporal Embedding module with two innovative GNNs for capturing complicated spatiotemporal interactions of heterogeneous urban phenomena and a Multivariate Probabilistic Prediction module for the prediction of both expected values and uncertainty. 
This framework can be adapted to various multivariate distributions based on different data characteristics. 
We conduct extensive experiments by comparing UQGNN with 12 state-of-the-art baselines in terms of 6 metrics using four real-world multivariate spatiotemporal datasets 
from Shenzhen, NYC, and Chicago. 
The experimental results show that our UQGNN effectively outperforms the baselines by improving the prediction accuracy and uncertainty quantification by 5\%. 


\begin{acks}
The authors would like to thank anonymous reviewers for their insightful comments and valuable suggestions. 
This work is partially supported by the National Science Foundation under Grant Nos. 2411152, 2401860, 2430700, and FSU Sustainability \& Climate Solutions Grant Program. Prof. Shenhao Wang and Prof. Yueheng Bu would like to thank the support from the Seed Grant at the University of Florida (ROSF2023). 
\end{acks}

\clearpage
\newpage
\balance
\bibliographystyle{ACM-Reference-Format}
\bibliography{reference}

\newpage
\appendix

\section{Framework Details}
\subsection{Adjacency Matrix}\label{adjacency}
The rationale for using the Gaussian threshold kernel is its ability to model spatial proximity and the smooth decay of interaction strength over distance since nearby nodes usually have stronger connections in urban networks. Also, the Gaussian kernel speeds up graph training by ignoring the weakest connections, which has been widely used in prior works~\cite{dcrnn, stzinb-gnn, stgcn}.
We utilize the z-score for normalization, which can be expressed as $(X-\mu)/\sigma$, where $\mu$ and $\sigma$ are the mean and the standard deviation of the original data. In addition, an offset is added to each dimension to avoid negative values because some distributions do not allow non-negative value characteristics, e.g., Poisson distribution and negative binomial distribution.

\subsection{Algorithm of MPP}\label{algo}
The number of regions is usually much more than the number of urban phenomena, which will make it more complex to handle.
For example, the Shenzhen dataset consists of 491 regions, so the dimension of $\bm{\Sigma}$ at each time step will be $491\times 491=241081$, which is infeasible to implement.
Meanwhile, we do not specifically pay attention to the correlation of the output time steps since they are not assumed to follow the distribution.
Our framework is not only applicable to the multivariate Gaussian distribution but also to other multivariate distributions. For instance, the multivariate Laplace distribution is characterized by two key parameters: $\mu$ (a vector representing the center of the distribution) and $\beta$ (a positive semi-definite matrix that describes the spread, orientation, and relationships between dimensions), so the algorithm can also handle this distribution. Although the negative binomial distribution consists of three key parameters including $r$ (number of successes), $p$ (success probability), $\mu$, and $\Sigma$ (correlations between the different dimensions), our UQGNN can also handle this by adding one more output block, with a procedure similar to the one described above. We also need to change the loss functions when adopting different multivariate distributions. Similarly, the loss function for a multivariate Laplace distribution can be expressed as:
\begin{equation}
    \begin{aligned}
\mathcal{L}(X;\bm{\mu},\bm{\beta}) &= \frac{1}{2} \log \left( (2\pi)^k |\beta| \right) + \frac{1}{2} (\mathbf{x} - \boldsymbol{\mu})^\top \beta^{-1} (\mathbf{x} - \boldsymbol{\mu})
    \end{aligned}
\end{equation}
The loss function for the multivariate negative binomial distribution can be written as:
\begin{equation}
    \begin{aligned}
\mathcal{L}(X; r, p,\bm{\mu}, \Sigma) &= - \sum_{i=1}^{M} \log \left( \frac{\Gamma(x_i + r_i)}{x_i! \Gamma(r_i)} \right) \\
&\quad + \sum_{i=1}^{M} \frac{r_i}{p_i} \log \left( 1 + \frac{p_i}{r_i} x_i \right) \\
&\quad + \frac{1}{2} \log |\Sigma| + \frac{1}{2} (\mathbf{x} - \boldsymbol{\mu})^\top \Sigma^{-1} (\mathbf{x} - \boldsymbol{\mu})
    \end{aligned}
\end{equation}


\section{Experiment Setup}\label{setup}
\subsection{Baselines}\label{baseli}
\subsubsection{\textbf{Deterministic Prediction Baselines}}
\begin{enumerate}[noitemsep, left=0pt]

\item STGCN\cite{stgcn}: Spatial-Temporal Graph Convolution Network combines spectral graph convolution with 1D convolution to capture spatial and temporal correlations.

\item DCRNN\cite{dcrnn}: Diffusion Convolutional Recurrent Neural Network integrates diffusion convolution with a sequence-to-sequence architecture to learn the representations of spatial dependencies and temporal relations.

\item GWNET\cite{GraphWavenet}: Graph WaveNet consists of a novel adaptive dependency matrix and a stacked dilated 1D convolution component, through which it can handle long sequences.

\item StemGNN\cite{cao2020spectral}: Spectral Temporal GNNs considers both intra-series temporal correlations and inter-series correlations simultaneously in multivariate time-series forecasting.

\item DSTAGNN\cite{dstagnn}: Dynamic Spatial-Temporal Aware Graph Neural Network can represent dynamic spatial relevance and acquire a wide range of dynamic temporal dependencies.

\item AGCRN\cite{bai2020adaptive}: AGCRN consists of a Node Adaptive Parameter Learning module to capture node-specific patterns and a Data Adaptive Graph Generation module to infer the inter-dependencies. 

\item SUMformer\cite{cheng2024rethinking}: As a Transformer-based method, SUMformer regards urban mobility as a complex multivariate time series. This perspective involves treating the time-varying values in each channel as individual time series.

\end{enumerate}

\subsubsection{\textbf{Probabilistic Prediction Baselines}}
\begin{enumerate}[noitemsep, left=0pt]

\item TimeGrad\cite{rasul2021autoregressive} is an auto-regressive model that combines the diffusion model with an RNN-based encoder.
\red{}

\item STZINB\cite{stzinb-gnn}: Spatial-Temporal Zero-Inflated Negative Binomial Graph Neural Network analyzes spatial and temporal correlations using diffusion and temporal convolution networks.

\item DiffSTG\cite{wen2023diffstg}: DiffSTG presents the first attempt to generalize the popular denoising diffusion probabilistic models to spatio-temporal graphs. 

\item DeepSTUQ\cite{uqunifi} combines the merits of variational inference and deep ensembling by integrating the Monte Carlo dropout and the Adaptive Weight Averaging re-training methods.

\item CF-CNN\cite{huang2024uncertainty}: The model proposes a conformalized GNN, extending conformal prediction to graph-based models. Given an entity in the graph, CF-GNN produces a prediction set/interval that provably contains the true label with pre-defined coverage probability.

\end{enumerate}

\subsection{Metrics}\label{allmetrics}
\subsubsection{\textbf{Deterministic Prediction Metrics}}\label{Deterministic Metrics}

We leverage three commonly used metrics, including Mean Absolute Error (MAE), Root Mean Squared Error (RMSE), and Mean Absolute Percentage Error (MAPE), to evaluate the performance of deterministic prediction. $Y$ is the ground truth label, and $\hat{Y}$ denotes the prediction result. $n$ is the number of predictions.
\begin{itemize}
\item Mean absolute error (MAE)
\begin{equation}
    \begin{aligned}
\text{MAE}=\frac{1}{n} \sum_{i=1}^{n}|Y_i-\hat{Y}_i|
    \end{aligned}
\end{equation}
\item Root mean squared error (RMSE)
\begin{equation}
    \begin{aligned}
    \text{RMSE} = \sqrt{\frac{1}{n}\sum_{i=1}^{n}(Y_i - \hat{Y}_i)^2}
    \end{aligned}
\end{equation}
\item Mean absolute percentage error (MAPE)
\begin{equation}
    \begin{aligned}
    \text{MAPE} = \frac{100\%}{n}\sum_{i=1}^{n}\left|\frac{Y_i - \hat{Y}_i}{Y_i}\right|
    \end{aligned}
\end{equation}

\end{itemize}

\subsubsection{\textbf{Probabilistic Prediction Metrics}}\label{Prob}

Three metrics including Continuous Ranked Probability Score (CRPS)~\cite{wen2023diffstg}, Kullback-Leibler Divergence (KL)~\cite{uqunifi}, and Mean Prediction Interval Width (MPIW)~\cite{stzinb-gnn} are used for uncertainty quantification evaluation.
\begin{itemize}[noitemsep, left=0pt]
\item Traditional metrics for accuracy evaluation such as MAE or RMSE are not directly applicable to probabilistic prediction. The Continuous Ranked Probability Score (CRPS) generalizes the MAE to the case of probabilistic prediction. The CPRS is one of the most widely used accuracy metrics for probabilistic prediction. which is used to measure the compatibility of an estimated probability distribution $F$ with an observation $x$:
\begin{equation}
    \begin{aligned}
    \text{CRPS}(F, x) = \int_{\mathbb{R}} \left( F(z) - \mathds{1}(z \geq x) \right)^2 \, \text{d}z
    \end{aligned}
\end{equation}
where $\mathds{1}(z \geq x)$ is the Heaviside step function (indicator function) which equals $1$ if $z \geq x$, and $0$ otherwise. Smaller CRPS means better performance.

\item Kullback-Leibler Divergence (KL-Divergence)
\begin{equation}
    \begin{aligned}
    \text{KL}=\sum_{i=1}^{N}Y_i\log \frac{Y_i}{\hat{Y}_i}
    \end{aligned}
\end{equation}
Smaller values are better since the KL-Divergence measures the difference between two distributions.

\item Mean Prediction Interval Width (MPIW)
\begin{equation}
    \begin{aligned}
    \text{MPIW}=\frac{1}{n} \sum_{i=1}^{N}(U(\hat{Y}_i)-L(\hat{Y}_i))
    \end{aligned}    
\end{equation}
where $U(\hat{Y}_i)$ and $L(\hat{Y}_i)$ represent the upper and lower bounds of prediction intervals corresponding to the $i^{th}$ region. Smaller prediction intervals are more desirable since they indicate high stability of the predicted values.
To be more specific, the MPIW of each mobility mode is $1.96\times \sigma$, where the $\sigma$ denotes the standard deviation of a specific mobility mode. 
This is because $95\%$ of the area under a Gaussian curve lies within approximately $1.96$ standard deviations of the mean and the overall MPIW value is the average of all mobility modes.

\end{itemize}

\subsection{Implementation Details}\label{datanormalization}
All experiments were conducted on a Linux platform equipped with an NVIDIA A100 GPU with 24 GB of memory.
During training, the Adam optimizer is employed with a batch size of 64. The initial learning rate is set to \(1 \times 10^{-3}\) and decays at a rate of \(5 \times 10^{-4}\) every 10 epochs. Early stopping is applied with a patience of 50 steps, based on the validation loss, to mitigate overfitting. 
The dataset is partitioned into training, validation, and testing subsets with a ratio of 8:1:1. The input sequence length is fixed at 12 time steps, and the prediction horizon is set to 1.

It is worth mentioning that traditional data normalization usually uses z-score, which can be expressed as $(X-\mu)/\sigma$, where $\mu$ and $\sigma$ are the mean and the standard deviation of the original data. 
The z-score method is very useful in deterministic prediction tasks, but it is not suitable for uncertainty quantification. This is because some distributions do not allow negative numbers, like the negative binomial distribution, but the z-score normalization will generate negative numbers, although all the original data are positive. 
Therefore, we utilize the max-min normalization instead.
In this way, we can normalize the data without negative normalized values.

\section{Computational Complexity Analysis}\label{appendix_complexity}

\noindent \textbf{\textit{Theoretical Analysis.}} The space and time complexities for the three key components in UQGNN are summarized in Table \ref{STcomplexity}, where $N$ is the number of nodes in the graph, $T$ is the time steps, $F$ is the number of urban phenomena, $D$ is total diffusion steps, and $M$ is the size of the eigenvector involved.
For MDGCN, the time complexity grows quadratically with $N$, due to the pairwise relationships (graph edges) being considered. This is acceptable in graph-based models, especially with sparsity optimizations, as real-world graphs often have sparse connectivity. 
ITCN is computationally efficient in both space and time dimensions, providing a strong foundation for handling temporal patterns.
For MPP, the cubic dependence on $M$ (calculating the eigenvalue) is manageable due to the limited range of $M$, while the linear dependence on $T$ and $N$ supports scalability for long sequences and large node sets. Hence, our UQGNN is theoretically computationally efficient for spatiotemporal prediction.

\begin{table}[h]
\caption{Theoretical analysis of our method's efficiency.} 
\centering
\begin{tabular}{c|cc}
\hline
Components    & Space Complexity   & Time Complexity\\ 
\hline
MDGCN & $\mathcal{O}(N \cdot T \cdot F \cdot D)$ & $\mathcal{O}(N^2 \cdot T \cdot F \cdot D)$  \\
ITCN & $\mathcal{O}(N \cdot T \cdot F)$ & $\mathcal{O}(N \cdot T \cdot F)$  \\
MPP & $\mathcal{O}(T \cdot N \cdot M^2)$ & $\mathcal{O}(N \cdot T \cdot M^3)$  \\
\hline
\end{tabular}\label{STcomplexity}
\vspace{-5pt}
\end{table}

\noindent \textbf{\textit{Empirical Analysis.}}
We also show the computational efficiency of our UQGNN empirically with model training time. The time to train one epoch on the four datasets using a batch size of 64 is shown in Table \ref{Tcomplexity}. We found our UQGNN has relatively short training times, which are comparable to the fastest baselines, highlighting the efficiency and scalability of our framework.

\begin{table}[h]
\setlength{\tabcolsep}{3pt}
\caption{Training time (seconds) for each epoch.} 
\centering
\begin{tabular}{c|cccc}
\hline

\multirow{2}*{Models} & \multicolumn{4}{c}{Datasets} \\
\cline{2-5}

    & Shenzhen   & NYC 1 & NYC 2 & Chicago\\ 
\hline
STGCN & 6.2 & 3.1 & 3.7 & 5.84 \\
DCRNN & 67.2 & 36.1 & 39.8 & 61.6 \\
GWNET & 16.1 & 8.3 & 10.7 & 13.9 \\
StemGNN & 15.3 & 7.8 & 10.1 & 14.4 \\
DSTAGCNN & 8.5 & 4.0 & 5.4 & 7.2 \\
AGCRN & 11.4 & 6.2 & 7.0 & 10.7 \\
SUMformer & 12.3 & 6.6 & 7.3 & 10.6 \\
TimeGrad & 8.8 & 4.7 & 5.6 & 7.9 \\
STZINB & 9.3 & 4.3 & 5.9 & 8.1 \\
DiffSTG & 9.5 & 4.6 & 5.7 & 8.3 \\
DeepSTUQ & 9.7 & 5.3 & 5.7 & 8.9 \\
CF-GNN & 10.1 & 4.6 & 6.3 & 9.5 \\
UQGNN & 8.7 & 4.4 & 5.2 & 8.2 \\
\hline
\end{tabular}\label{Tcomplexity}
\end{table}


\end{document}